# Task-Semantic Graph-Driven Distributed Agent Networking for Underwater Target Tracking

Shengchao ZHU[1], Guangjie HAN[2*], Chuan LIN[3] & Yu He[2]

1 College of Computer Science and Software Engineering, Hohai University, Nanjing 210098, China
2 College of Information Science and Engineering, Hohai University, Changzhou 210013, China
3 Software College, Northeastern University, Shenyang 110819, China

**Abstract** Autonomous underwater vehicle (AUV) swarms are emerging as intelligent underwater networks, where each node must sense, communicate, process local data, and make decisions under severe acoustic constraints. Persistent underwater target tracking is a representative task in this setting. The target moves over time, the communication topology changes as the AUVs move, acoustic links are intermittent and bandwidth-limited, and each AUV can observe only partial target and neighbor information. Multi-agent reinforcement learning (MARL) is a natural candidate for distributed tracking, yet existing studies still lack a unified open-source platform for evaluating different MARL algorithms under six-degree-of-freedom AUV dynamics. In addition, policies trained directly from raw geometric states and low-level force/torque actions often struggle to represent task phases, observation reliability, link quality, and local cooperation roles. This paper addresses these issues by developing an open-source MARL-AUV platform that integrates DI-engine with a six-degree-of-freedom underwater AUV target-tracking simulator. To the best of our knowledge, it is the first open platform that connects a public MARL training framework with physically modeled AUV swarm-based tasks, and provides a unified experimental protocol for fair training, testing, and comparison of representative RL and MARL algorithms. Based on this platform, we propose STG-MAPPO, a **S**emantic **T**ask **G**raph-enhanced variant of **M**ulti-**A**gent **P**roximal **P**olicy **O**ptimization, to support task-semantic distributed agent networking. STG-MAPPO builds semantic policy inputs from tracking diagnostics, task phases, observation confidence, link availability, neighbor tracking quality, and local role advantage. A compact semantic task graph links communication-constrained network states to decentralized actor decisions, and a velocity-level action abstraction maps high-level cooperative decisions to executable six-degree-of-freedom AUV control inputs. Extensive experiments demonstrate that STG-MAPPO consistently outperforms mainstream MARL baselines and conventional AUV control MARL settings in persistent tracking accuracy, target-loss rate, convergence stability, and action smoothness under both nominal and stressed conditions. The code is available at https://github.com/dasjsaj/MARL-AUV.

## 1 Introduction

Next-generation communication networks are moving beyond connection-centric infrastructure toward task-oriented intelligent systems. In these networks, unmanned platforms, underwater sensors, edge nodes, and autonomous robots are no longer simple data transmitters. They also sense the environment, process local data, perform computation, and make decisions. This shift makes distributed autonomous agent networking a central problem, especially in harsh environments where global information exchange is impractical and cooperation must be built from local observations and limited communication resources [1-4].

Underwater AUV swarm target tracking is a typical example. Underwater acoustic communication is limited by narrow bandwidth, long propagation delay, time-varying link quality, and intermittent connectivity [5]. At the same time, AUV motion follows nonlinear six-degree-of-freedom dynamics and is constrained by attitude stability, propulsion limits, and energy consumption. Persistent target tracking therefore goes beyond single-vehicle trajectory control. Multiple AUVs must keep the target observable, exchange useful neighbor information, and adjust their cooperation as the target moves and the communication topology changes [6].

MARL is a natural candidate for learning distributed cooperative policies from local observations [7]. The centralized-training and decentralized-execution paradigm is particularly relevant, as it allows richer information to be used for value estimation during training while keeping each agent locally deployable at execution time [8]. However, applying MARL to underwater AUV networks remains difficult in practice. Existing MARL implementations are often built for abstract benchmarks or simplified robotic systems, and thus do not provide a unified interface for fair comparison under physically constrained AUV dynamics [9]. In addition, raw geometric observations and low-level force/torque actions are not sufficient for stable underwater tracking. A policy must recognize whether an AUV is searching, approaching, stably tracking, losing the target, or reacquiring it. It must also assess the reliability of neighbor information from link quality, observation confidence, and tracking error [10]. Without such task-semantic information, cooperation has to be inferred implicitly from continuous states, which makes learning harder and may lead to unstable control.

* Corresponding author(hanguangjie@gmail.com)

To address the above issues, we integrate DI-engine [11] with a high-fidelity six-degree-of-freedom AUV target-tracking simulator and develop an open-source MARL-AUV platform. The platform provides a unified interface for training, testing, logging, and comparing multiple representative MARL algorithms under the same AUV dynamics, task settings, stress conditions, and evaluation metrics. To the best of our knowledge, MARL-AUV is the first open-source platform that connects a public MARL training framework with physically modeled AUV swarm target-tracking environments, enabling reproducible and one-command evaluation of MARL algorithms for underwater AUV tasks. The platform is available at https://github.com/dasjsaj/MARL-AUV.

Built on the MARL-AUV platform, we examine the difficulty of applying MARL to nonlinear AUV control. A policy must simultaneously capture tracking context, observation reliability, neighbor value, and six-degree-of-freedom physical constraints, which makes it difficult to learn stably from raw states and force/torque actions. This difficulty further leads to slow convergence and unstable policy optimization. To address this, we propose STG-MAPPO, namely Semantic Task Graph-enhanced Multi-Agent Proximal Policy Optimization. It extracts task-level semantics from tracking diagnostics, task phases, observation quality, link availability, neighbor tracking error, and local role advantage, and organizes them into a compact semantic task graph for decentralized decision-making. A velocity-level action interface is further used to map high-level cooperative decisions to executable AUV control inputs. This design offers a task-semantic way to apply MARL to complex nonlinear physical systems. The main contributions are summarized as follows.

- First, we develop an open-source MARL-AUV benchmarking platform by integrating DI-engine with a six-degree-of-freedom AUV target-tracking simulator. The platform provides a unified interface for training, testing, logging, and comparing representative MARL algorithms in underwater AUV tasks.
- Second, we formulate AUV swarm target tracking as a communication-constrained distributed agent networking problem. This formulation encodes task phase, observation quality, link availability, neighbor tracking error, and local role advantage into interpretable semantic states.
- Third, we propose STG-MAPPO, a Semantic Task Graph-enhanced MAPPO framework under the centralized-training and decentralized-execution paradigm. It combines semantic task graph modeling with MAPPO, enabling each AUV to make decentralized decisions from local semantic observations.
- Finally, we introduce a velocity-level execution interface compatible with six-degree-of-freedom AUV dynamics, and evaluate the proposed method against representative MARL baselines under many conditions.

# 2 Related work

## 2.1 Underwater AUV network-based cooperative target tracking

Underwater multi-AUV networks have become a key platform for long-term ocean observation, cooperative search, and mobile target tracking. Unlike terrestrial or aerial multi-agent systems, they operate under narrow-band acoustic communication, long propagation delays, intermittent connectivity, limited sensing range, and strongly constrained vehicle dynamics. These factors make cooperative tracking different from centralized trajectory control. In such settings, AUVs must not only estimate and follow a moving target, but also sustain useful local interactions as communication and observation conditions change over time.

Prior work has studied this problem from several angles. Hierarchical software-defined multi-AUV reinforcement learning introduces networking and MARL into underwater target tracking, while interrupted software-defined multi-AUV reinforcement learning further considers time-saving policy learning under intermittent execution conditions [12, 13]. Model-based tracking control has also been explored. For example, trajectory-prediction-based AUV tracking combines target prediction with control design to track dynamic underwater targets [14]. Other studies incorporate multi-source sensing and multi-AUV cooperation. Acoustic-optical fusion compensates for delayed and low-frequency acoustic measurements with visual observations when tracking highly maneuvering underwater targets [15], and bearing-only multi-AUV tracking addresses cooperative estimation with limited relative information [16]. Learning-from-demonstration and cooperative localization studies further suggest that multi-AUV tracking and localization depend on both interaction modeling and physically feasible motion execution [17, 18].

These studies provide important foundations, but most of them emphasize tracking accuracy, localization quality, policy convergence, or sensor fusion. The networking relation among AUVs is often treated as a given communication condition, a software-defined infrastructure, or an implicit part of the MARL state. In communication-constrained target tracking, however, the usefulness of a neighbor depends on link availability, observation confidence, tracking error, and its role in the current task phase. This calls for a task-semantic view of underwater cooperative tracking, in which AUV cooperation is guided not only by proximity or fixed topology, but also by the task value of locally available information.

### 2.2 Task-oriented agent networking and semantic state representation

Task-oriented communication and semantic representation offer a relevant lens for studying agent networking. Rather than sending all raw observations, a task-oriented system seeks to retain the information that matters for downstream decision-making or inference. Recent work on deep learning-based semantic communication has shown that semantic features can be learned and transmitted to support task performance under communication constraints [19, 20]. Related studies further show that semantic transmission should account for interpretability, information importance, and feature selection, since different pieces of information contribute differently to a given task [21-23]. Similar ideas also appear in intelligent task-oriented semantic communication studies published in the Journal on Communications, where communication effectiveness is evaluated by task utility instead of bit-level reconstruction accuracy [24,25].

For distributed autonomous agents, however, task-oriented networking involves more than compressing sensory data. Each agent must decide which local and neighboring information is useful for its current action. In cooperative MARL, graph-based and information-aware methods have been used to improve strategy evaluation and decision modeling in large-scale or partially observable multi-agent systems [26]. These studies indicate that structured relational information can ease the difficulty of learning cooperation directly from raw joint states. However, most existing work on semantic communication and MARL is not designed for underwater AUV networks, where the state representation must jointly describe sensing reliability, acoustic link availability, neighbor tracking quality, and physically feasible control.

Our method follows this task-oriented view but targets a different system setting. Instead of treating semantic representation as a standalone communication codec, it builds a semantic task graph for distributed AUV networking. The graph summarizes task phase, observation quality, link condition, neighbor tracking error, and local role advantage into a compact policy input. In this formulation, semantic information is not an auxiliary description of the environment. It serves as an intermediate representation that connects communication-constrained network states with decentralized policy decisions.

## 3 System model and problem formulation

### 3.1 Underwater intelligent AUV network model

Consider an underwater intelligent agent network composed of $N$ AUVs, denoted by $\mathcal{A}=\{1,2,\ldots,N\}$. At discrete time $t$, the position, velocity, and attitude of the $i$-th AUV are denoted by $p_i^t\in\mathbb{R}^3$, $v_i^t\in\mathbb{R}^3$, and $\eta_i^t$, respectively. The position and velocity of the moving target are denoted by $p_g^t\in\mathbb{R}^3$ and $v_g^t\in\mathbb{R}^3$. Each AUV acts as a sensing node, a communication node, and a local decision-making node. Due to acoustic link constraints, target-motion uncertainty, and local sensing range limitations, each agent can only access its own state, target-relative observations, neighbor-relative states, and local task-semantic features.

The distance between AUV $i$ and the target, together with the tracking error, is defined as

$$d_i^t=\| p_i^t-p_g^t \|_2,\qquad e_i^t=|d_i^t-d^*|. \tag{1}$$

where $d_i^t$ is the Euclidean distance between AUV $i$ and the target, $d^*$ is the desired target observation distance, and $e_i^t$ is the distance-based tracking error.

### 3.2 Six-degree-of-freedom AUV dynamics

To capture underwater physical constraints, a six-degree-of-freedom AUV dynamic model is adopted. Let the generalized pose be $\eta=[x,y,z,\phi,\theta,\psi]^{\mathrm{T}}$, the body-frame velocity be $\nu=[u,v,w,p,q,r]^{\mathrm{T}}$, and the generalized force/torque input be $\tau=[X,Y,Z,K,M,N]^{\mathrm{T}}$. The continuous-time kinematics and dynamics can be written as

$$\dot{\eta}=J(\eta)\nu, \tag{2}$$

$$M\dot{\nu}+C(\nu)\nu+D(\nu)\nu+g(\eta)=\tau+w. \tag{3}$$

In Eq. (2), $J(\eta)$ is the transformation matrix from the body frame to the inertial frame. In Eq. (3), $M$ is the mass and added-mass matrix, $C(\nu)$ is the Coriolis and centripetal matrix, $D(\nu)$ is the hydrodynamic damping matrix, $g(\eta)$ is the vector of gravity, buoyancy, and restoring moments, and $w$ denotes environmental disturbances. The RL policy does not alter the dynamics directly. Instead, at each decision step, it outputs an action that is converted into an executable control input $\tau$ through the environment interface, and the six-degree-of-freedom dynamics then propagate the AUV state.

## 3.3 Communication-constrained local observation model

AUVs mainly rely on acoustic links for information exchange. Due to propagation delay, bandwidth limitation, distance attenuation, and environmental noise, it is difficult to maintain global, real-time, and highly reliable communication among all agents. The local communication relation of the AUV swarm at time $t$ is represented by a dynamic graph

$$\mathcal{G}_c^t = (\mathcal{A}, \mathcal{E}_c^t), \tag{4}$$

where $\mathcal{A}$ is the AUV node set and $\mathcal{E}_c^t$ is the available communication edge set. For any two nodes $i$ and $j$, the normalized link availability is defined as

$$q_{ij}^t = \mathrm{clip}\left(1 - \frac{\| p_j^t - p_i^t \|_2}{R_c}, 0,1\right), \tag{5}$$

where $R_c$ is the effective communication or proximity radius, and $\mathrm{clip}(\cdot)$ restricts the link value to $[0,1]$. If $q_{ij}^t > 0$, node $j$ is considered an available neighbor of node $i$. This distance-dependent model captures the dynamic nature of local communication relations caused by AUV mobility.

To further characterize acoustic-link uncertainty, packet loss, propagation delay, and information freshness are incorporated into an extended link-quality model:

$$q_{ij}^t = \exp\left(-\alpha d_{ij}^t\right)\left(1 - p_{\mathrm{loss},ij}^t\right)\exp\left(-\frac{\Delta t_{ij}}{\tau_f}\right), \tag{6}$$

where $d_{ij}^t = \| p_j^t - p_i^t \|_2$, $\alpha$ is a distance attenuation coefficient, $p_{\mathrm{loss},ij}^t$ is the packet loss probability, $\Delta t_{ij}$ is the age of neighbor information, and $\tau_f$ is the freshness decay constant.

## 3.4 Task state modeling for AUV swarm networks

In underwater AUV swarm target tracking, communication, sensing, data, and intelligent decision-making jointly affect task completion quality. Communication states determine whether neighbor information is available, sensing states determine whether target observations are reliable, data states reflect historical tracking errors and neighbor tracking quality, and intelligence states indicate policy inference and action selection based on local information. To incorporate these factors into distributed policy learning, task-related resources are represented as a local semantic state vector:

$$z_i^t = \left[z_{\mathrm{sen},i}^t, z_{\mathrm{com},i}^t, z_{\mathrm{data},i}^t, z_{\mathrm{int},i}^t\right]. \tag{7}$$

where $z_{\mathrm{sen},i}^t$ includes target distance, observation confidence, and target-loss indicators; $z_{\mathrm{com},i}^t$ includes the effective-neighbor ratio, mean link quality, and best link quality; $z_{\mathrm{data},i}^t$ includes neighbor tracking errors, moving averages of historical errors, and information freshness; and $z_{\mathrm{int},i}^t$ includes task phase, observation quality, and local role advantage. This modeling turns task-related sensing, communication, data, and intelligence factors into a unified state representation for semantic task graph construction.

## 3.5 Dec-POMDP formulation

Under local observation and communication constraints, AUV swarm target tracking is formulated as a decentralized partially observable Markov decision process (Dec-POMDP):

$$\mathcal{M} = \langle \mathcal{S}, \{\mathcal{O}_i\}_{i=1}^N, \{\mathcal{A}_i\}_{i=1}^N, P, R, \gamma, N \rangle. \tag{8}$$

where $\mathcal{S}$ is the global state space, $\mathcal{O}_i$ is the local observation space of AUV $i$, $\mathcal{A}_i$ is its action space, $P$ is the state transition function, $R$ is the shared or mixed reward function, and $\gamma \in (0,1)$ is the discount factor.

Additionally, the centralized-training and decentralized-execution (CTDE) paradigm is adopted. During training, the critic may use global or concatenated state information to estimate values, while during execution, each actor generates actions only from local observations:

$$\pi_\theta(a^t | o^t) = \prod_{i=1}^{N} \pi_\theta\,(a_i^t | o_i^t), \tag{9}$$

$$J(\theta) = \mathbb{E}_{\pi_\theta}\left[\sum_{t=0}^{T-1} \gamma^t R^t\right]. \tag{10}$$

The objective is to learn a distributed policy under communication constraints and partial observations, so that the AUV swarm can reduce long-term tracking error, target-loss rate, and control oscillation while preserving a deployable local decision-making form.

# 4 Semantic task graph driven distributed agent networking method

This section presents a semantic task graph-driven distributed agent networking method for communication-constrained underwater AUV swarm target tracking. The proposed method formulates target tracking as a task-oriented autonomous networking process rather than a pure motion control problem. To realize this idea, we integrate semantic task graph modeling with multi-agent proximal policy optimization (MAPPO) [27] and present Semantic Task Graph-enhanced MAPPO (STG-MAPPO). It organizes sensing, communication, data, and decision-related information into task-semantic policy inputs, thereby enabling each AUV to make decentralized decisions under partial observation and constrained underwater communication.

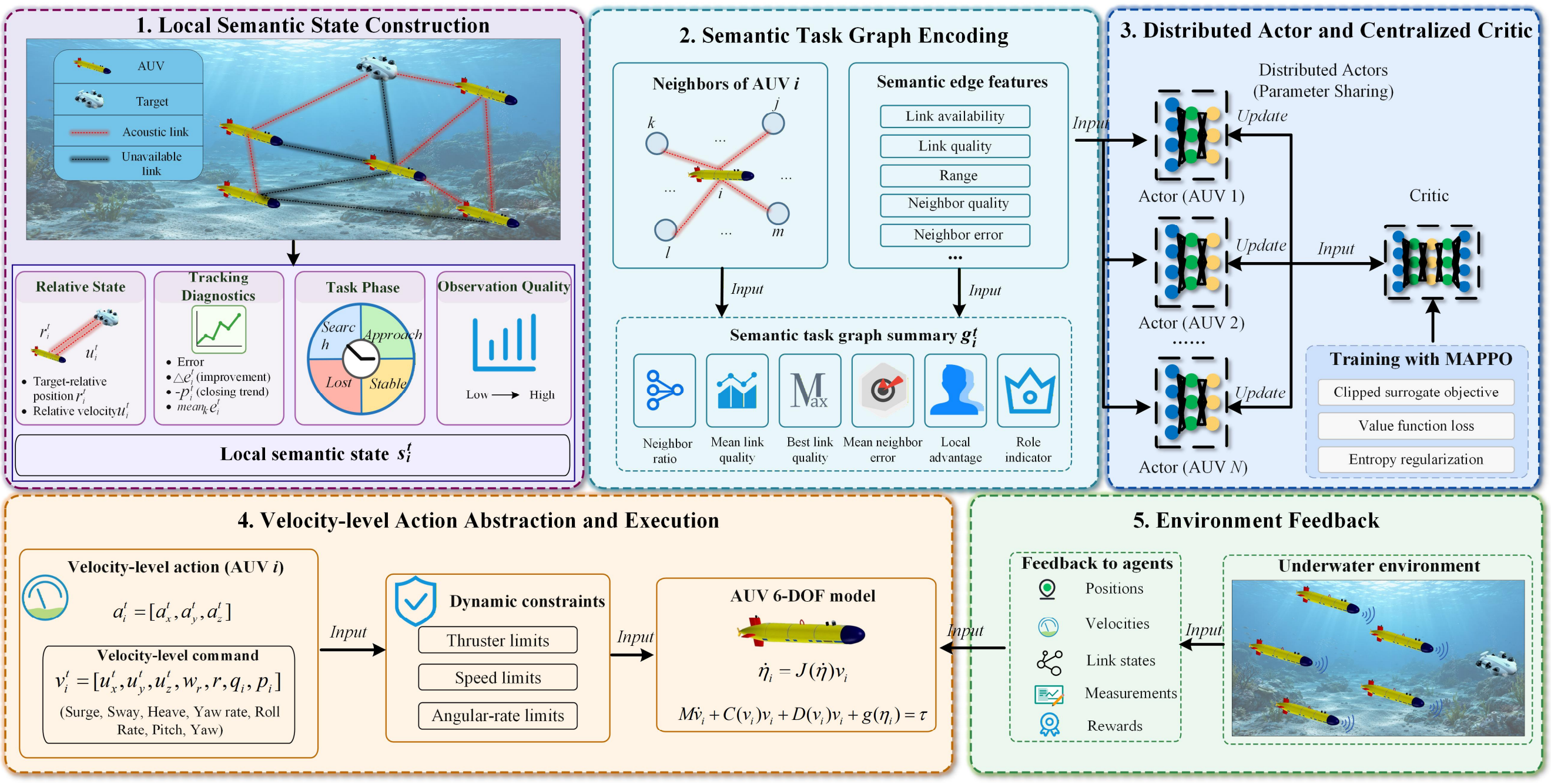

**Figure 1** Overall framework of STG-MAPPO.

## 4.1 Design rationale and overall framework

Conventional MARL methods usually pass positions, velocities, target distances, and neighbor-relative states directly to the policy network, so the cooperation structure among agents has to be learned implicitly. This setting can be adequate for simple tasks, but it is less suitable for underwater target tracking. The policy must behave differently when an AUV is searching for the target, moving toward it, maintaining a stable tracking state, or recovering from target loss. At the same time, acoustic links vary with distance and environmental conditions, and direct exploration in the raw six-degree-of-freedom force and torque space can lead to action saturation and slow convergence. STG-MAPPO therefore introduces task semantics and communication-aware neighbor summaries before action selection.

The semantic task graph is used as an intermediate representation between communication-constrained network states and distributed decision-making, rather than as an auxiliary reward term. It combines link availability, target observation confidence, tracking-error trends, and local role indicators into the actor input. This design allows each AUV to make decisions based on both its own tracking state and the task value of neighboring information, supporting autonomous networking and cooperative control for persistent target tracking.

## 4.2 Local semantic state construction

For distributed execution, relative states are preferred over global coordinates. The target-relative position and velocity of AUV $i$ are defined as

$$r_i^t = p_g^t - p_i^t, \qquad u_i^t = v_g^t - v_i^t. \tag{11}$$

Here, $r_i^t$ describes the target direction relative to AUV $i$, and $u_i^t$ describes relative motion. Relative states are more suitable for dynamic networking and reduce the influence of global coordinate scaling.

To make tracking quality directly observable to the policy, the error-improvement and closing-trend variables are defined as

$$\Delta e_i^t = e_i^{t-1} - e_i^t, \qquad \rho_i^t = d_i^t - d_i^{t-1}. \tag{12}$$

A positive $\Delta e_i^t$ indicates that the tracking error decreases at the current step, while a negative $\rho_i^t$ indicates that the AUV is approaching the target. The observation confidence is computed from the target distance and sensor range:

$$c_i^t = \text{clip}\left(1 - \frac{d_i^t}{R_s}, 0, 1\right), \tag{13}$$

where $R_s$ is the effective sensing range. The tracking diagnostic vector is then defined as

$$z_{\text{track},i}^t = \left[\bar{d}_i^t, \bar{e}_i^t, \Delta\bar{e}_i^t, -\bar{\rho}_i^t, \text{mean}_k(e_i), c_i^t, l_i^t, \ell_i^t/T\right]. \tag{14}$$

In (14), barred variables are normalized values, $\text{mean}_k(e_i)$ is the moving average of tracking errors over the latest $k$ steps, $l_i^t$ is the target-loss indicator, $\ell_i^t$ is the consecutive loss duration, and $T$ is the maximum episode length. This vector transforms tracking distance, error trend, observation confidence, and loss state into policy-ready semantic features.

## 4.3 Task phase and observation-quality semantics

Target tracking exhibits clear task phases. An AUV must increase observation opportunities during searching, reduce target distance during approaching, maintain the desired distance band during stable tracking, and reacquire the target after loss. The task phase is defined as

$$\text{phase}_i^t \in \{\text{search}, \text{approach}, \text{stable}, \text{lost}\}. \tag{15}$$

The phase is determined by the target-loss indicator, target distance, and error-improvement term. If $l_i^t = 1$, the node is in the lost phase. If the target is not lost and $d_i^t$ is within the near range, it is in the stable phase. If the target is not lost and the error improvement is positive, it is in the approach phase. The remaining cases are assigned to the search phase. The observation-quality semantic variable is defined as

$$\text{quality}_i^t \in \{\text{high}, \text{medium}, \text{low}, \text{unavailable}\}. \tag{16}$$

It is determined by thresholds on $c_i^t$. Both the phase and observation-quality variables are rule-based and physically interpretable rather than external labels or additional learned classifiers. They are encoded as one-hot vectors and used together with the tracking diagnostic vector.

## 4.4 Semantic task graph encoding and autonomous networking representation

In a communication-constrained underwater environment, cooperation relations should not be treated as fixed topology. They vary with node positions, link quality, target-observation states, and neighbor tracking capability. The AUV swarm at time $t$ is represented by a semantic task graph

$$\mathcal{G}_s^t = (\mathcal{V}, \mathcal{E}^t, \mathcal{Z}^t), \tag{17}$$

where $\mathcal{V}$ is the AUV node set, $\mathcal{E}^t$ is the edge set formed by local communication or proximity relations, and $\mathcal{Z}^t$ contains node and edge semantic features. To reduce communication overhead, each node constructs a compact semantic graph summary instead of transmitting the full graph. For AUV $i$, the semantic task graph feature is defined as

$$g_i^t = \left[\frac{|\mathcal{N}_i^t|}{N}, \text{mean}_{j\in\mathcal{N}_i^t}(q_{ij}^t), \max_{j\in\mathcal{N}_i^t}(q_{ij}^t), \text{mean}_{j\in\mathcal{N}_i^t}(e_j^t), \text{adv}_i^t, \text{role}_i^t\right]. \tag{18}$$

The local advantage term is defined as

$$\text{adv}_i^t = \text{clip}\left(\text{mean}_{j\in\mathcal{N}_i^t}(e_j^t) - e_i^t, -1, 1\right). \tag{19}$$

If $\text{adv}_i^t > 0$ and $c_i^t$ exceeds the observation-confidence threshold, AUV $i$ is assigned a local primary-tracking role indicator. This role is not a fixed identity but changes with target position, observation quality, and neighbor states. The final policy input is

$$o_i^{\text{STG},t} = \left[o_{i,\text{raw}}^t, z_{\text{track},i}^t, z_{\text{sem},i}^t, g_i^t\right], \tag{20}$$

where $o_{i,\text{raw}}^t$ is the physical observation, $z_{\text{track},i}^t$ is the tracking diagnostic vector, $z_{\text{sem},i}^t$ is the task-phase and observation-quality encoding, and $g_i^t$ is the semantic task graph summary.

## 4.5 Distributed policy decision based on semantic task graphs

With (20), the distributed actor of each AUV is expressed as

$$a_i^t \sim \pi_\theta\left(\cdot \mid o_i^{\text{STG},t}\right). \tag{21}$$

During execution, AUV $i$ uses only its own observation and neighbor summary. It neither accesses the global state nor relies on a centralized controller. When a node has high observation confidence, low tracking error, and reliable links, its local primary-tracking tendency increases. When its observation quality degrades or the target is lost, the policy can adjust its motion using neighbor tracking errors and link qualities. Thus, cooperation is generated through task-semantic evaluation rather than predefined roles.

This mechanism reflects task-driven autonomous networking. Nodes do not need to be manually assigned as leaders, followers, or relays; instead, they estimate their cooperative value from local and neighbor semantic states. Since the exchanged information consists of low-dimensional states such as position, velocity, observation confidence, tracking error, and link quality, the communication load is compatible with low-bandwidth acoustic links.

## 4.6 Velocity-level action abstraction for six-degree-of-freedom AUV execution

If the actor directly outputs a six-dimensional force/torque vector, the policy must simultaneously learn cooperative tracking and low-level control in a strongly coupled dynamic space. STG-MAPPO therefore adopts a three-dimensional velocity action as the task-level decision interface:

$$a_i^t = \left[a_x, a_y, a_z\right] \in [-1,1]^3. \tag{22}$$

The desired velocity is

$$v_{\text{des},i}^t = s_v a_i^t, \tag{23}$$

where $s_v$ is the velocity scale. The environment then generates six-degree-of-freedom control inputs from the desired velocity and the current velocity. The force component is

$$\tau_{\text{force},i}^t = \text{clip}\left(K_v R(\eta_i^t)^{\text{T}}\left(v_{\text{des},i}^t - v_i^t\right), -\tau_{\max}, \tau_{\max}\right), \tag{24}$$

and the moment component for attitude stabilization and angular damping is

$$\tau_{\text{moment},i}^t = \text{clip}\left(-K_\eta[\phi_i^t, \theta_i^t, 0]^{\text{T}} - K_\omega[p_i^t, q_i^t, r_i^t]^{\text{T}}, -\tau_{m,\max}, \tau_{m,\max}\right). \tag{25}$$

The final control input is

$$\tau_i^t = \left[\tau_{\text{force},i}^t, \tau_{\text{moment},i}^t\right]. \tag{26}$$

Before execution, the action is clipped, rate-limited, and smoothed:

$$a_{\text{smooth}}^t = (1-\alpha)a_{\text{delta}}^t + \alpha a^{t-1}. \tag{27}$$

This abstraction does not remove the six-degree-of-freedom dynamics. Instead, it restricts the policy output to task-level velocity decisions and lets the physical execution layer complete force/torque mapping, thereby preserving dynamic realism while reducing exploration difficulty.

## 4.7 Semantic reward for persistent target tracking

The reward function is designed for persistent target tracking rather than encirclement or formation closure. Let the stable tracking distance band be $[d_l, d_u]$. The distance-band deviation is

$$b_i^t = \max(0, d_l - d_i^t, d_i^t - d_u), \tag{28}$$

and the corresponding band-stability score is

$$B_i^t = \text{clip}\left(1 - \frac{b_i^t}{e_{\max}}, 0,1\right). \tag{29}$$

The tracking reward of AUV $i$ is

$$r_{\text{track},i}^t = \text{clip}\big(w_b B_i^t + w_p P_i^t + w_c C_i^t + w_r R_i^t - w_u U_i^t, -1,1\big), \tag{30}$$

where $P_i^t$ denotes error improvement, $C_i^t$ denotes closing tendency, $R_i^t$ denotes reacquisition benefit, and $U_i^t$ denotes a too-close penalty. The observation reward is

$$r_{\text{obs},i}^t = \text{clip}(c_i^t - l_i^t, -1,1). \tag{31}$$

The coordination and communication rewards are

$$r_{\text{coord}}^t = \text{clip}\left(1 - \frac{\text{mean}_i(b_i^t)}{e_{\max}}, 0,1\right), \tag{32}$$

$$r_{\text{comm}}^t = \text{mean}_{(i,j)\in\mathcal{E}^t} q_{ij}^t. \tag{33}$$

The semantic phase reward adapts the optimization emphasis according to the current task phase:

$$r_{\text{sem},i}^t = \begin{cases} P_i^t, & \text{phase}_i^t = \text{lost}, \\ 0.65P_i^t + 0.35C_i^t - 0.25U_i^t, & \text{phase}_i^t \in \{\text{search}, \text{approach}\}, \\ 0.55c_i^t + 0.45(1 - U_i^t), & \text{phase}_i^t = \text{stable}. \end{cases} \tag{34}$$

The action cost is

$$\text{cost}_i^t = \text{clip}(0.6 \parallel a_i^t \parallel_2^2 + 0.4 \parallel a_i^t - a_i^{t-1} \parallel_2^2, 0,1). \tag{35}$$

The global reward is then

$$R^t = w_T r_{\text{track}}^t + w_O r_{\text{obs}}^t + w_C r_{\text{coord}}^t + w_M r_{\text{comm}}^t + w_S r_{\text{sem}}^t - w_A \text{cost}^t. \tag{36}$$

The tracking component carries the dominant weight, while semantic, communication, and coordination terms serve as auxiliary structures. Thus, the objective is not to maximize link quality itself, but to use communication and semantic information to support persistent target tracking.

## 4.8 STG-MAPPO optimization

STG-MAPPO follows the centralized-training and decentralized-execution paradigm. The joint policy is

$$\pi_\theta(a^t|o^t) = \prod_{i=1}^{N} \pi_\theta\left(a_i^t|o_i^{\text{STG},t}\right). \tag{37}$$

The objective is to maximize the discounted cumulative reward:

$$J(\theta) = \mathbb{E}_{\pi_\theta}\left[\sum_{t=0}^{T-1} \gamma^t R^t\right]. \tag{38}$$

The temporal-difference residual and generalized advantage estimator are

$$\delta_t = R^t + \gamma V_\phi(s^{t+1}) - V_\phi(s^t), \tag{39}$$

$$A_t = \sum_{l=0}^{T-t-1} (\gamma\lambda)^l \delta_{t+l}. \tag{40}$$

The PPO clipped actor objective is

$$L_{\text{actor}}(\theta) = \mathbb{E}_t[\min(r_t(\theta)A_t, \text{clip}(r_t(\theta), 1-\epsilon, 1+\epsilon)A_t) + \beta H(\pi_\theta)], \tag{41}$$

where

$$r_t(\theta) = \frac{\pi_\theta(a_t|o_t)}{\pi_{\theta_{\text{old}}}(a_t|o_t)}. \tag{42}$$

The critic loss and overall optimization objective are

$$L_{\text{critic}}(\phi) = \mathbb{E}_t\left[\left(V_\phi(s_t) - V_t^{\text{target}}\right)^2\right], \tag{43}$$

$$L(\theta, \phi) = -L_{\text{actor}}(\theta) + c_v L_{\text{critic}}(\phi). \quad (44)$$

Compared with standard MAPPO, the main differences lie not in the optimizer but in the semantic actor input, the velocity-level action interface, and the task-oriented reward structure.

# 5 Evaluations

This section evaluates the proposed STG-MAPPO framework in a six-degree-of-freedom AUV target-tracking simulator integrated with the DI-engine multi-agent training interface [11]. The experiments are designed to assess whether semantic task graph modeling, communication-aware state construction, semantic rewards, and velocity-level action abstraction jointly improve persistent target tracking under decentralized execution.

## 5.1 Environment and training configuration

Experiments are conducted in a six-degree-of-freedom AUV target-tracking simulator integrated with the DI-engine multi-agent training interface. The environment contains four AUVs and one moving target. In this paper, we do not further study dynamic changes in the number of AUVs, since this issue has been addressed in our previous IEEE TMC work [28], where a lightweight population-adaptation mechanism was developed to adjust the AUV team size after a single training process. Each episode lasts 500 steps, and all methods are trained for 2.0e6 environment steps with evaluations every 5000 steps. We consider both medium and hard scenarios, where the medium scenario serves as the main benchmark and the hard scenario introduces more challenging target motion and observation conditions. All results are averaged over three random seeds, and final-stage statistics are computed over the last 20% of evaluation checkpoints.

We compare STG-MAPPO with representative MARL baselines, including MAPPO [27], HAPPO [29], MASAC [30], MADDPG [31], MADQN [32], and MATD3 [33] baselines. STG-MAPPO uses semantic-enhanced observations, semantic task graph summaries, semantic tracking rewards, and the velocity-level action interface, whereas the baselines use non-semantic observations and their corresponding action settings. The evaluation focuses on both task performance and execution stability, including evaluation return, mean tracking error, tail target distance, target lost rate, action saturation rate, and control cost. Distance-related metrics are reported in kilometers. This protocol is designed to assess whether the proposed semantic task graph and velocity-level execution mechanism improve persistent target tracking rather than only increasing cumulative reward. The detailed parameters are provided in the Appendix.

## 5.2 Main comparison in the medium scenario

In the medium scenario, we evaluate all trained methods under two test settings: the nominal setting and the combined-stress setting. The nominal setting follows the standard evaluation configuration, while the combined-stress setting increases target maneuverability, enlarges the initial AUV-target separation, and degrades sensing and communication conditions. For each method, we reload the best checkpoint obtained during medium-scenario training and report the results over three random seeds, with thirty evaluation episodes for each seed. Table 1 summarizes the average target distance, target lost rate, and evaluation return under the two settings.

Table 1 Performance comparison in the medium scenario under nominal and combined-stress test settings.

| Algorithm | Nominal dist. (km) | Combined dist. (km) | Nominal lost | Combined lost | Avg. return |
|---|---|---|---|---|---|
| STG-MAPPO | **0.012 ± 0.001** | **0.013 ± 0.002** | **0.000 ± 0.000** | **0.000 ± 0.000** | **1018.84 ± 15.93** |
| MAPPO | 0.185 ± 0.182 | 0.221 ± 0.199 | 0.053 ± 0.091 | 0.121 ± 0.194 | 720.53 ± 299.58 |
| MADDPG | 0.175 ± 0.258 | 0.256 ± 0.391 | 0.048 ± 0.083 | 0.178 ± 0.309 | 756.38 ± 366.89 |
| MATD3 | 0.594 ± 0.216 | 0.965 ± 0.100 | 0.440 ± 0.184 | 0.646 ± 0.083 | 200.58 ± 239.84 |
| HAPPO | 0.421 ± 0.040 | 0.713 ± 0.045 | 0.019 ± 0.032 | 0.617 ± 0.076 | 267.96 ± 190.40 |
| MADQN | 0.680 ± 0.092 | 0.749 ± 0.108 | 0.593 ± 0.153 | 0.611 ± 0.100 | 79.80 ± 73.67 |
| MASAC | 1.172 ± 0.002 | 1.219 ± 0.038 | 0.938 ± 0.054 | 0.875 ± 0.042 | -256.10 ± 39.26 |

Table 1 compares the best medium-trained checkpoints under the nominal and combined-stress test settings. STG-MAPPO maintains an average target distance of only 0.012 km in the nominal setting and 0.013 km under combined stress, with zero target loss in both cases. By contrast, MAPPO and MADDPG, the two strongest non-STG baselines in this setting, show noticeably larger distances under combined stress, reaching 0.221 km and 0.256 km, respectively, and their target-loss rates also increase. Other baselines degrade more severely, especially in target distance and loss probability. These results suggest that STG-MAPPO learns a more stable tracking policy rather than a policy that only performs well under standard evaluation conditions. The semantic task graph helps the agents use observation quality, link availability, and neighbor tracking states more effectively, while the velocity-level action interface supports persistent tracking without relying on unstable low-level force/torque exploration.

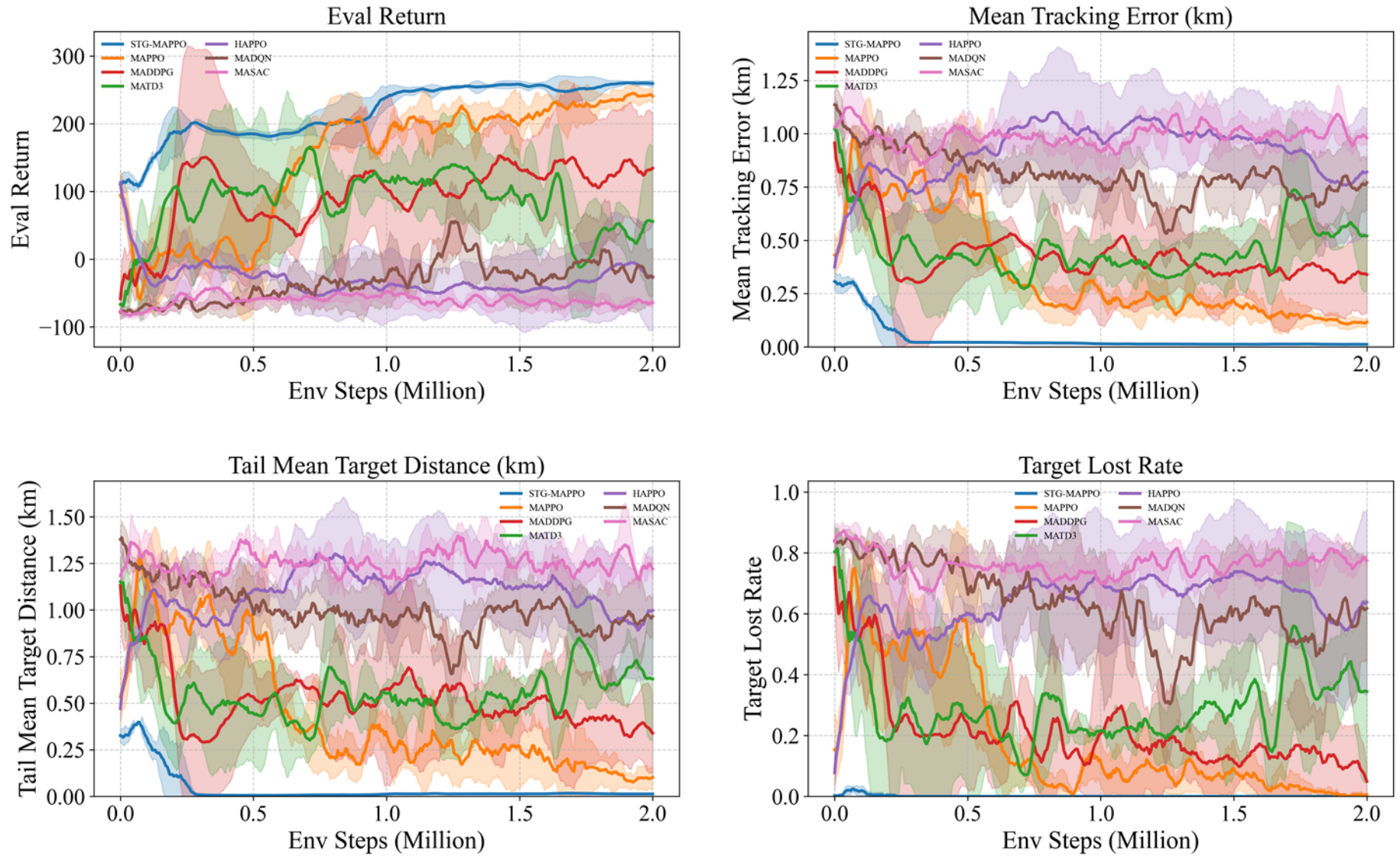

Figure 2 Convergence curves in the medium 4-AUV scenario.

Figure 2 reports the training dynamics in the medium scenario from four perspectives: evaluation return, mean tracking error, tail target distance, and target lost rate. STG-MAPPO shows a consistent learning pattern across these metrics. Its return increases quickly and then remains stable, while both the tracking error and tail target distance decrease to a low level after the early training stage. More importantly, the target lost rate is reduced to nearly zero and does not rebound in later evaluations, indicating that the policy learns to maintain the target over time rather than only reducing the distance temporarily. In comparison, the baseline methods either converge more slowly or show larger fluctuations across seeds. Several of them also retain nonzero target-loss rates in the late stage, which suggests that their learned policies are less reliable for persistent tracking. These curves show that the advantage of STG-MAPPO lies not only in the final performance, but also in faster convergence and more stable late-stage behavior.

Figure 3 summarizes the final-stage performance in the medium scenario, where each metric is averaged over the last 20% of evaluation checkpoints. STG-MAPPO achieves the best overall result. It attains the highest return while keeping the tail target distance, tracking error, and target lost rate at the lowest levels. This is important because a high return alone does not necessarily imply stable tracking; the policy must also keep the target within an effective observation range and avoid late-stage loss. MAPPO performs better than most other baselines, but its remaining tracking error and tail distance are still much larger than those of STG-MAPPO. The comparison indicates that the improvement mainly comes from the task-semantic policy representation and the velocity-level action interface, which make the tracking objective easier to learn than using raw non-semantic observations or low-level control actions.

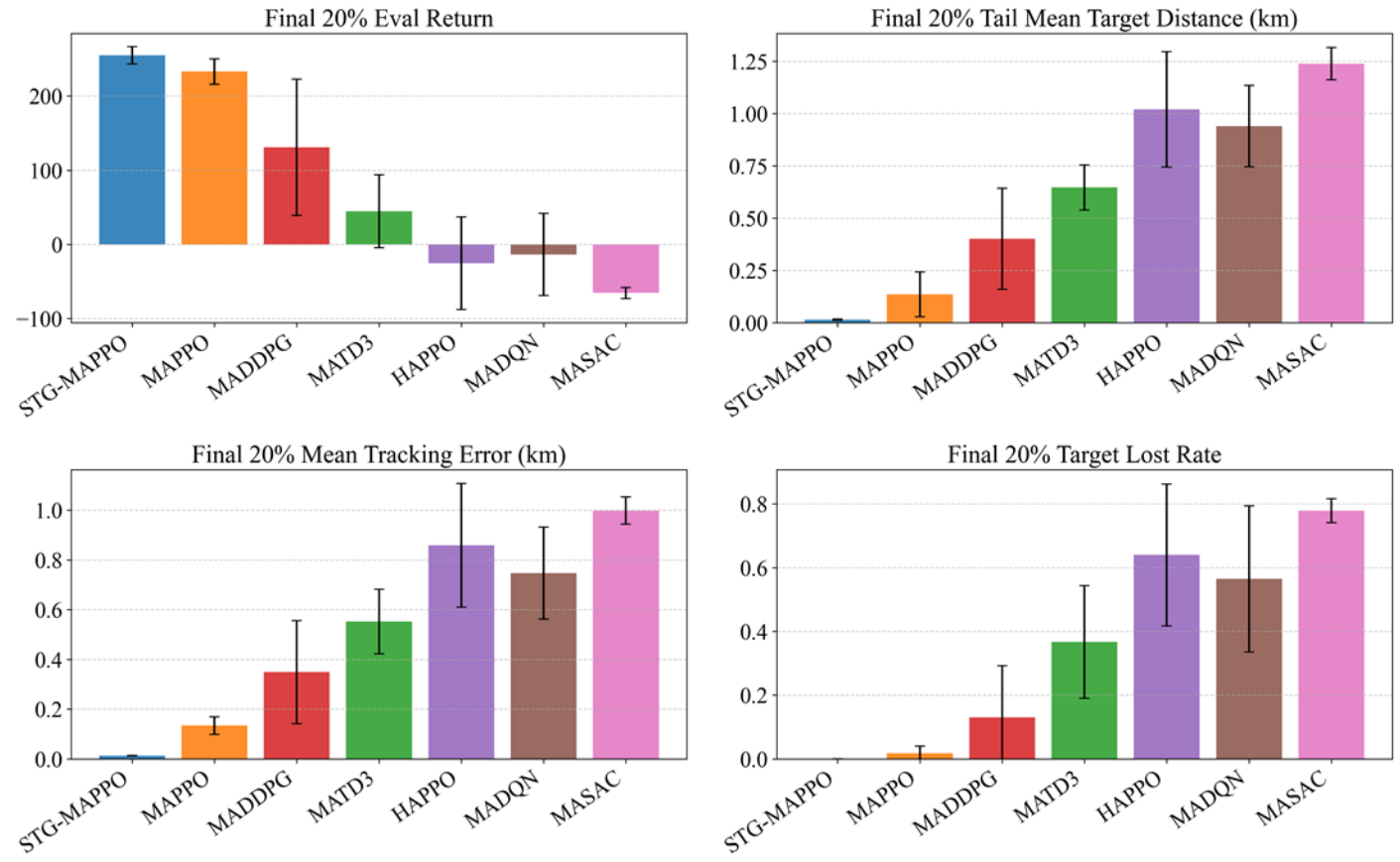

Figure 3 Final-stage comparison in the medium 4-AUV scenario over the last 20 percent of evaluation checkpoints.

## 5.3 Robustness analysis in the hard scenario

In the hard scenario, we further examine the robustness of each method under more challenging tracking conditions. Compared with the medium scenario, the target is harder to maintain within the sensing range, and useful neighbor information becomes less stable due to degraded observation and communication conditions. This setting therefore tests whether a learned policy can still identify the current tracking phase, use reliable neighbor cues, and execute stable control when the task becomes less favorable. We evaluate the hard-trained checkpoints under both nominal and combined-stress settings, and report the corresponding target distance, target lost rate, and evaluation return in Table 2.

Table 2 Performance comparison in the hard scenario under nominal and combined-stress test settings.

| Algorithm | Nominal dist. (km) | Combined dist. (km) | Nominal lost | Combined lost | Avg. return |
|---|---|---|---|---|---|
| STG-MAPPO | **0.013 ± 0.002** | **0.013 ± 0.004** | **0.000 ± 0.000** | **0.000 ± 0.000** | **981.71 ± 33.50** |
| MAPPO | 0.064 ± 0.009 | 0.095 ± 0.040 | 0.000 ± 0.000 | 0.002 ± 0.003 | 899.77 ± 68.37 |
| MADDPG | 0.346 ± 0.399 | 0.282 ± 0.308 | 0.187 ± 0.324 | 0.138 ± 0.234 | 613.34 ± 298.83 |
| MATD3 | 0.554 ± 0.378 | 0.584 ± 0.386 | 0.332 ± 0.292 | 0.308 ± 0.297 | 275.03 ± 279.69 |
| HAPPO | 0.518 ± 0.256 | 0.674 ± 0.377 | 0.361 ± 0.288 | 0.576 ± 0.428 | 218.68 ± 282.49 |
| MADQN | 0.735 ± 0.253 | 0.812 ± 0.296 | 0.514 ± 0.229 | 0.499 ± 0.232 | 36.53 ± 169.35 |
| MASAC | 1.200 ± 0.117 | 1.254 ± 0.130 | 0.839 ± 0.010 | 0.881 ± 0.083 | -274.77 ± 28.11 |

Table 2 reports the best-checkpoint results in the hard scenario under both nominal and combined-stress settings. STG-MAPPO keeps the average target distance at 0.013 km in both settings and achieves zero target loss, showing that its tracking behavior remains stable even when the task difficulty increases. MAPPO is the closest baseline, with a low nominal target-loss rate and a combined-stress distance of 0.095 km, but it still shows a clear gap from STG-MAPPO in distance accuracy. The remaining baselines degrade more noticeably: MADDPG and MATD3 show larger variance across seeds, while HAPPO, MADQN, and MASAC suffer from higher target-loss rates under stress. These results indicate that the semantic task graph helps the policy maintain reliable tracking cues when observations and neighbor information become less stable, rather than only improving performance under easier evaluation conditions.

Figure 4 presents the convergence behavior in the hard scenario. STG-MAPPO maintains a stable learning trajectory across all four metrics: the evaluation return increases steadily, while the mean tracking error, tail target distance, and target lost rate remain consistently low after convergence. In particular, the target lost rate drops to zero and does not show late-stage rebound, which indicates that the learned policy can sustain target visibility under more difficult tracking conditions. In contrast, most baselines either converge more slowly or exhibit clear oscillations in distance-related metrics. This suggests that raw geometric observations alone are not sufficient for stable cooperation when both observation quality and communication reliability degrade.

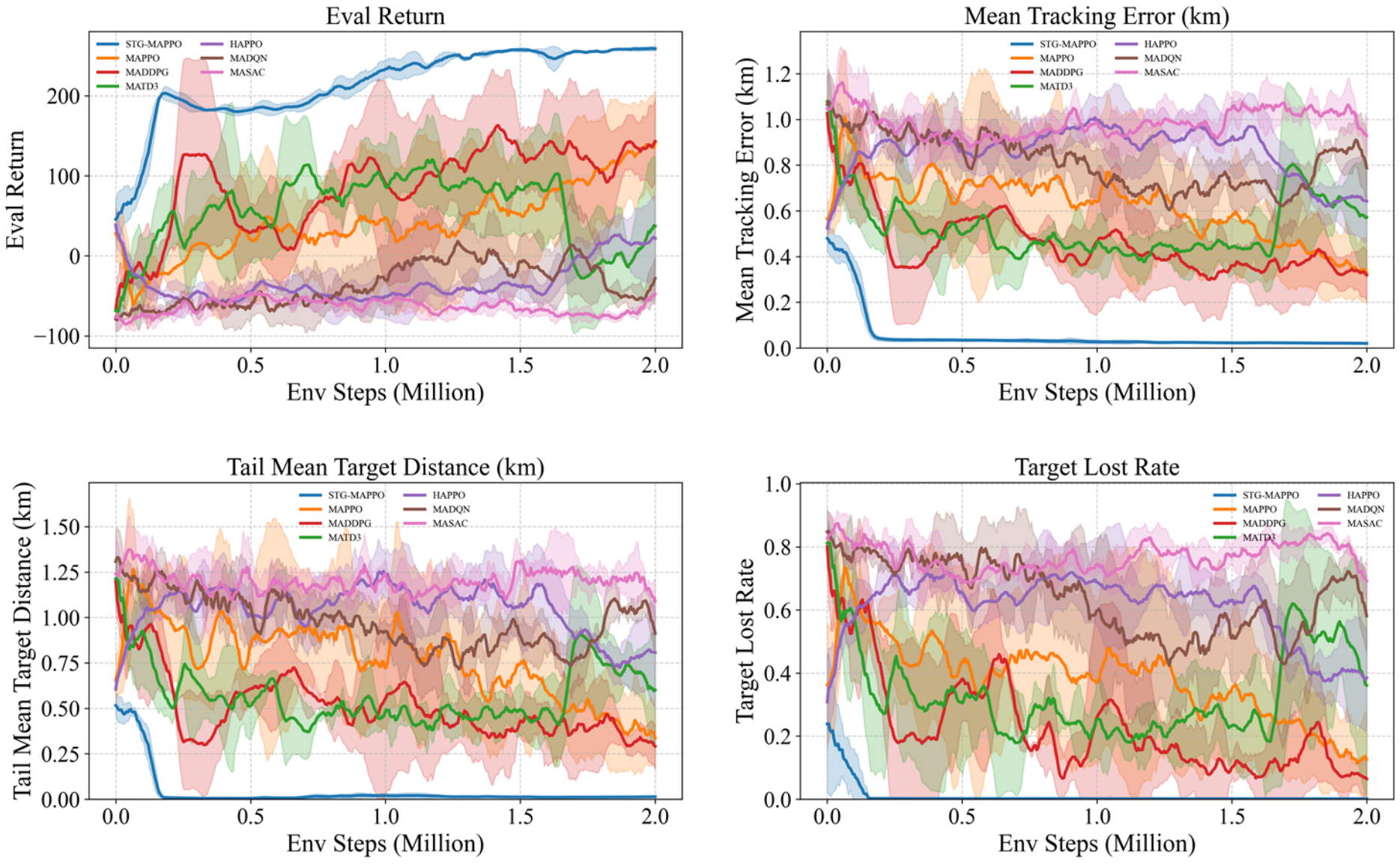

Figure 4 Convergence curves in the hard 4-AUV scenario.

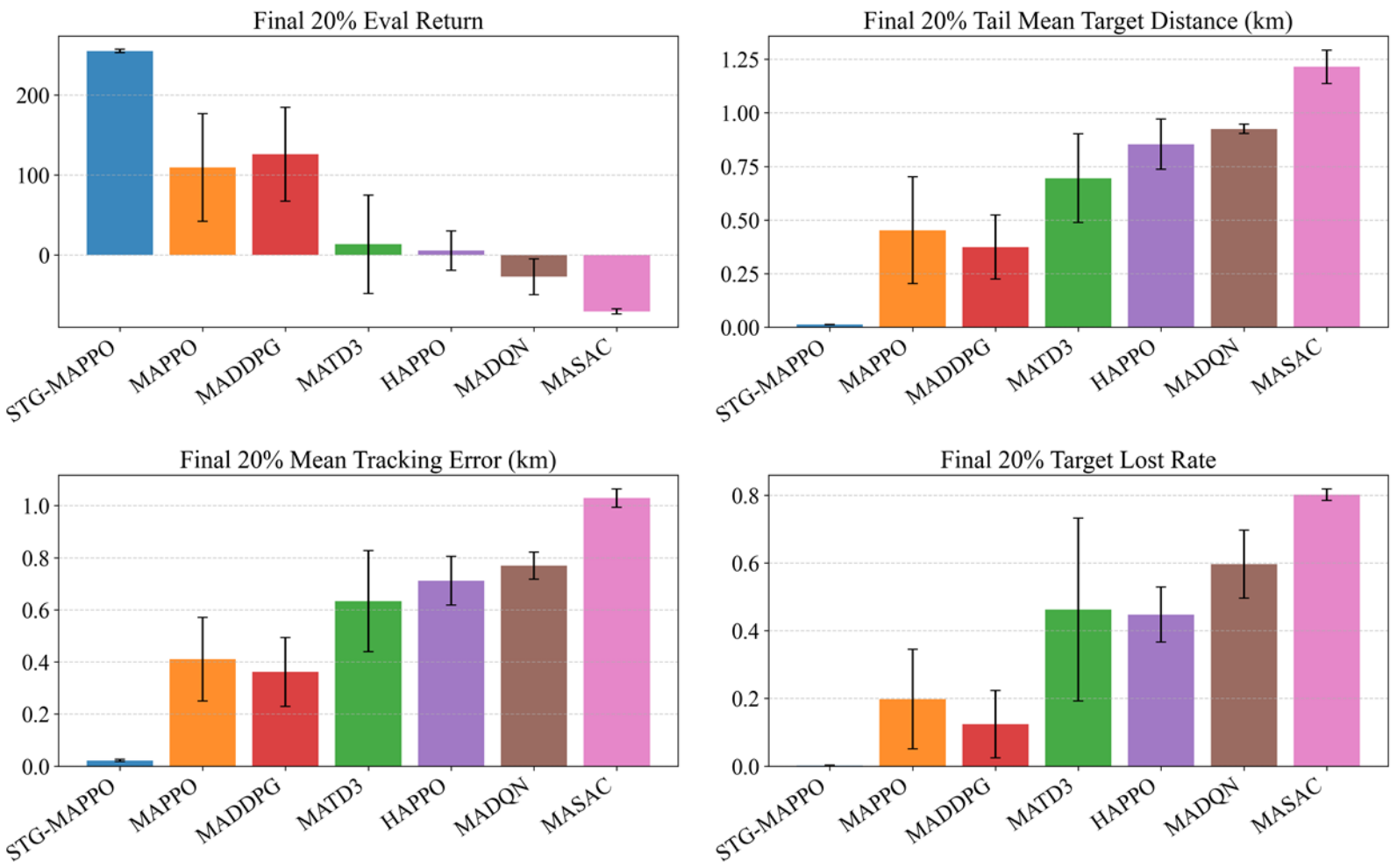

Figure 5 Final-stage comparison in the hard 4-AUV scenario over the last 20 percent of evaluation checkpoints.

Figure 5 reports the final-stage performance in the hard scenario, averaged over the last 20% of evaluation checkpoints. STG-MAPPO achieves the highest return while keeping the tail target distance, tracking error, and target lost rate at the lowest levels. This result is important because the hard scenario places stronger demands on both observation reliability and neighbor cooperation. Most non-semantic baselines either maintain larger distance errors or still suffer from target-loss events after training, suggesting that their policies are less robust when useful observations and communication links become unstable. By contrast, STG-MAPPO benefits from explicit task-phase, observation-confidence, link-quality, and local-role cues, which help the actors make more reliable decentralized decisions under degraded underwater conditions.

## 5.4 Test for the tracking trajectories

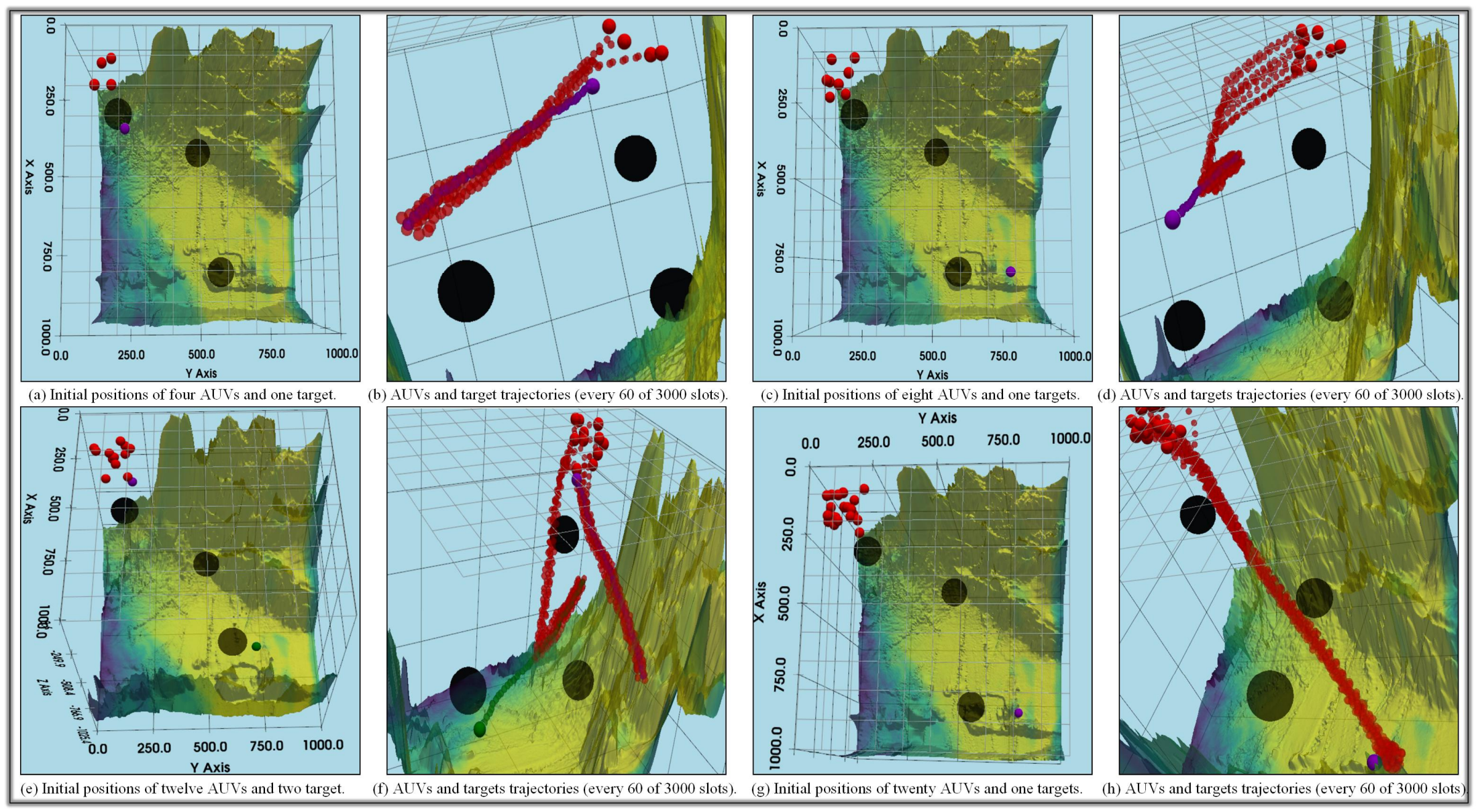

Figure 6 GEBCO-based qualitative visualization of STG-MAPPO in realistic underwater terrain. Red spheres denote AUVs, purple and green spheres denote targets, and black spheres denote obstacles. Figures (a)–(b), (c)–(d), (e)–(f), and (g)–(h) correspond to 4-AUV/1-target, 8-AUV/1-target, 12-AUV/2-target, and 20-AUV/1-target settings, respectively.

To further evaluate the applicability of STG-MAPPO in realistic underwater terrain, we conduct qualitative visualization experiments using GEBCO bathymetric data from 21.2°N to 22.5°N and 121.6°E to 122.9°E. We first test the basic setting with four AUVs tracking one target. Then, based on our previously published IEEE TMC work on incremental MARL-based AUV population adaptation [28], we extend the swarm size to 8, 12, and 20 AUVs. Figure 6 presents four representative cases. Figures 6(a) and 6(b) show the initial positions and tracking trajectories of four AUVs tracking one target. Figures 6(c) and 6(d) show eight AUVs tracking one target. Figures 6(e) and 6(f) show twelve AUVs tracking two targets. Figures 6(g) and 6(h) show twenty AUVs tracking one target. Red spheres denote AUVs, purple and green spheres denote moving targets, and black spheres denote obstacles. The trajectories are sampled every 60 time slots over 3000 time slots.The visualization results show that STG-MAPPO can maintain intelligent, stable, and scalable underwater target-tracking behaviors over realistic seabed terrain under different swarm sizes and target configurations.

## 5.5 Ablation experiments

In this subsection, we conduct ablation experiments to examine the contribution of each major component in STG-MAPPO. Four variants are considered. MAPPO-raw-tau6 directly learns policies in the six-degree-of-freedom force/torque action space. MAPPO-velocity3 introduces the velocity-level action abstraction but does not use semantic task graph modeling. MAPPO-semantic-state further incorporates semantic state information, while removing the full semantic reward and graph-driven decision loop. STG-MAPPO-full denotes the complete proposed method. This comparison allows us to distinguish the effect of action abstraction from that of semantic state construction and semantic task graph modeling.

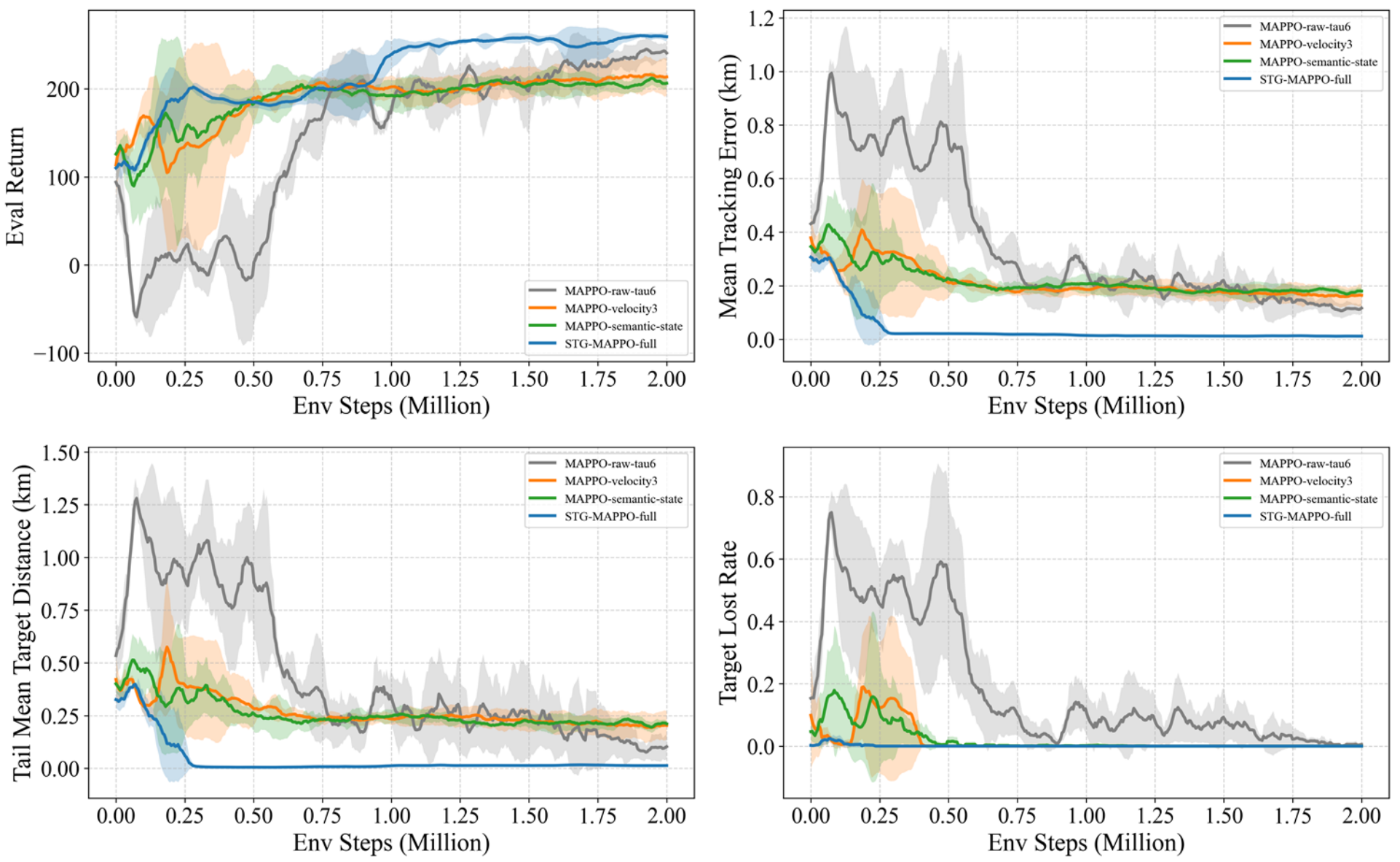

Figure 7 Ablation convergence curves.

Figure 7 provides a detailed ablation view of how different components affect learning. The raw force/torque variant, MAPPO-raw-tau6, shows large oscillations in the early stage: its return drops sharply, while the tracking error, tail target distance, and target lost rate all increase substantially. This indicates that directly exploring in the coupled six-degree-of-freedom action space makes policy learning unstable. After replacing the raw action with the velocity-level interface, MAPPO-velocity3 becomes much more stable and quickly suppresses target-loss events, showing that action abstraction effectively reduces the difficulty of low-level control. However, its tracking error and tail distance remain at a relatively high level, which suggests that action abstraction alone is insufficient for precise persistent tracking. MAPPO-semantic-state further improves stability by introducing task-related semantic information, but its final distance-related errors are still close to those of MAPPO-velocity3. In contrast, STG-MAPPO-full rapidly converges within the early training stage, keeps the target lost rate at zero, and reduces both mean tracking error and tail target distance to near-zero levels. These results show that the three components play complementary roles: velocity-level actions stabilize physical execution, semantic states provide task context, and the full semantic task graph with semantic reward guides the agents toward more accurate and persistent cooperative tracking.

Figure 8 reports the final-stage ablation results averaged over the last 20% of evaluation checkpoints. STG-MAPPO-full achieves the best overall performance: it obtains the highest evaluation return while reducing the tail mean target distance to a near-zero level. This indicates that the complete design not only improves cumulative reward but also maintains the target within a stable tracking range at the end of each episode. MAPPO-velocity3 and MAPPO-semantic-state both eliminate target loss, but their tail target distances remain around 0.2 km, showing that action abstraction or semantic states alone can improve stability but are insufficient for precise persistent tracking. MAPPO-raw-tau6 obtains a relatively high return and a smaller mean tail distance than the two partial variants, but it exhibits a nonzero target lost rate with large variance, which suggests that direct force/torque exploration can occasionally approach the target but does not guarantee reliable target maintenance. The action saturation rate is close to zero for all variants, indicating that the superiority of STG-MAPPO-full does not come from more aggressive boundary actions. Instead, the improvement is mainly due to the joint effect of velocity-level execution, semantic state construction, and semantic task graph-guided optimization, which together enable accurate and stable underwater target tracking.

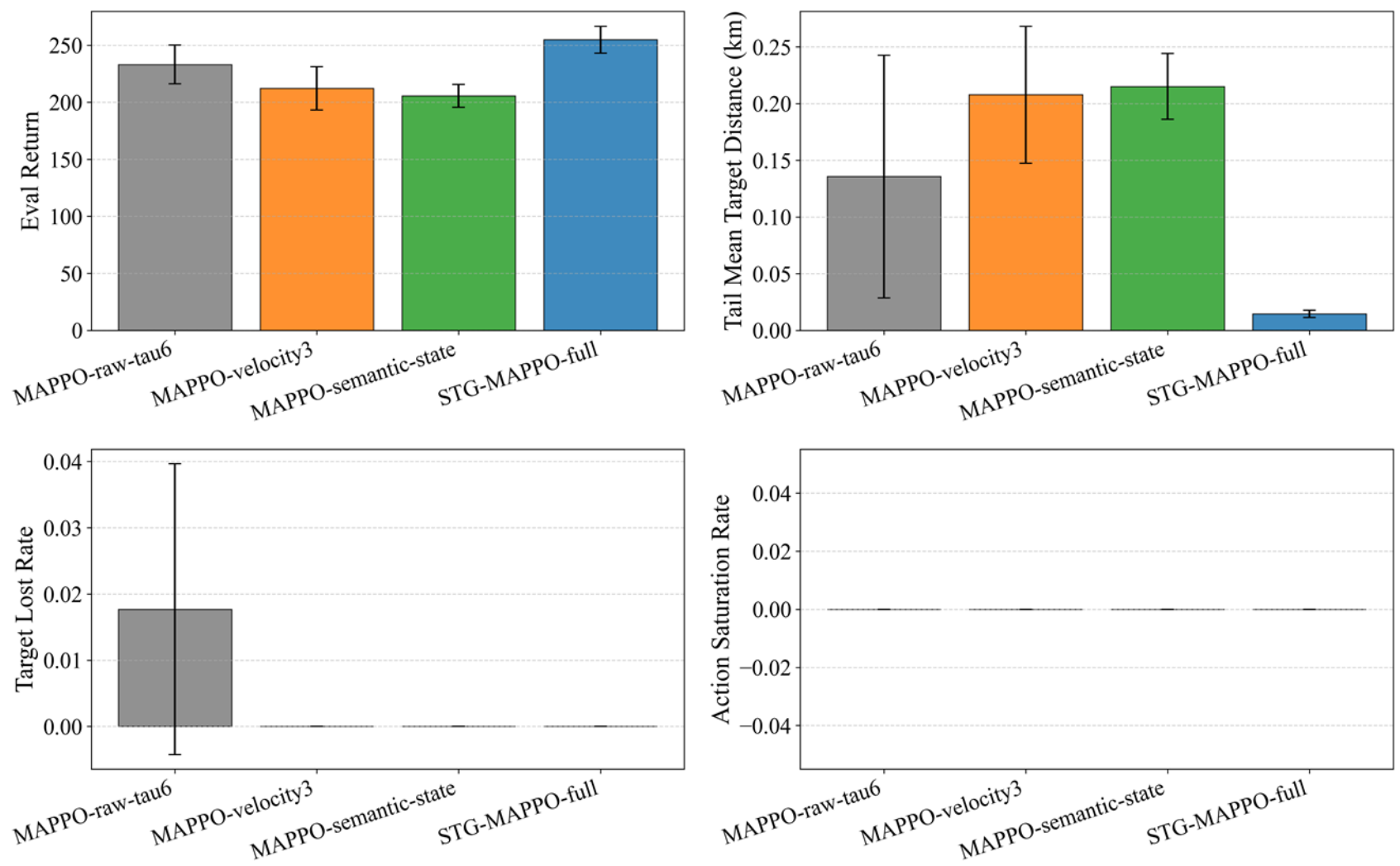

Figure 8 Final-stage ablation comparison over the last 20 percent of evaluation checkpoints.

# 6 Conclusion

This paper investigated distributed autonomous agent networking for underwater AUV swarm target tracking under constrained acoustic communication. We first developed an open-source MARL-AUV platform that integrates DI-engine with a six-degree-of-freedom AUV target-tracking simulator, providing a unified interface for training, testing, and comparing representative MARL algorithms in physically constrained underwater tasks. On this basis, we proposed STG-MAPPO, a semantic task graph-enhanced MARL framework that connects communication-constrained network states with decentralized policy decisions. By encoding tracking diagnostics, task phases, observation confidence, link availability, neighbor tracking quality, and local role advantage into semantic-enhanced observations, STG-MAPPO enables each AUV to reason about both its own tracking state and the task value of neighboring information. The velocity-level action abstraction further reduces the difficulty of direct exploration in the coupled six-degree-of-freedom force/torque space, while the component-wise semantic reward guides the policy toward persistent tracking rather than transient target approaching. Experiments in medium, hard, and stress-test scenarios show that STG-MAPPO achieves more stable convergence, lower tracking error, lower target-loss rate, and smoother control than representative MARL baselines. Future work will extend the platform with more realistic acoustic-channel models, explicit information-freshness constraints, heterogeneous AUV/USV collaboration, and field-test validation in physical underwater environments.

**Acknowledgements** The work is supported by the Joint Funds of the National Natural Science Foundation of China under Grant No. 62572169, 62572107 and the Open Fund of State Key Laboratory of Acoustics and Marine Information under Grant SKLA202501.

# Supplementary Appendix “Task-Semantic Graph-Driven Distributed Agent Networking for Underwater Target Tracking”

This standalone appendix supplements the main manuscript with complete metric definitions, additional figures, full stress-test tables, ablation results, and pseudocode. It is intended to be submitted as supplementary material or used as an internal experiment record.

## Appendix Structure

A. Metric definitions: explains the tracking, stability, control, and robustness metrics.

B. Medium-scenario supplementary experiments: includes additional figures and complete final-stage tables.

C. Hard-scenario supplementary experiments: includes corresponding hard-scenario figures and tables.

D. Best-checkpoint stress tests: reports all perturbation conditions and all key indicators.

E. Ablation supplementary experiments: provides ablation figures and complete tables.

F. Pseudocode: summarizes STG-MAPPO training and checkpoint stress-test evaluation.

## A. Metric Definitions

**Table S1 Metric definitions used in supplementary experiments.**

| Metric | Meaning |
|---|---|
| Eval return | Episode-level cumulative evaluation return. Higher is better. |
| Tail mean target distance (km) | Mean Euclidean distance between all AUVs and the target over the final evaluation tail. Lower is better. |
| Mean tracking error (km) | Mean absolute deviation between the AUV-target distance and the desired tracking distance. Lower is better. |
| Target lost rate | Fraction of evaluation steps in which the target is outside effective observation/tracking conditions. Lower is better. |
| Action saturation rate | Fraction of actions near clipping or actuator saturation. Lower values indicate smoother and less aggressive control. |
| Control cost | Penalty associated with action magnitude, action variation, and motion effort. Lower is preferable when tracking accuracy is similar. |
| Tracking reward | Reward component directly associated with distance reduction, tracking progress, and maintaining effective observation. |
| Semantic reward | Reward component generated from task phase, observation quality, communication quality, and role semantics. |
| Communication quality | Diagnostic indicator of effective local observation/communication availability. Higher is better. |

All final-stage statistics in the medium, hard, and ablation tables are reported as mean +/- standard deviation across seeds unless otherwise noted. For best-checkpoint stress tests, values are computed over independent evaluation episodes under each perturbation condition.

**Table S2 Evaluation parameters.**

| Notation | Description | Value |
|---|---|---|
| | **General simulation and evaluation settings** | |
| N | Number of AUV agents | 4 |
| Target | Number of targets | 1 |
| T | Episode length / evaluation horizon | 500 |
| Training budget | Maximum training budget per run | $2.0 \times 10^6$ |
| Time step | Simulation time step | 0.1 s |
| Workspace | Three-dimensional workspace boundary | $x, y, z \in [-1.0, 1.0]$ |
| Seeds | Random seeds | $\{0, 1, 2\}$ |
| Evaluation interval | Evaluation interval | 5000 |
| **AUV six-degree-of-freedom dynamic parameters** | | |
| Model | AUV dynamic profile | REMUS100 MSS-style 6DOF model |
| M | Mass and added-mass model | REMUS100 MSS-style parameters |
| m | Vehicle mass | 31.9 kg |
| Inertia | Rigid-body rotational inertia | $I_x = 0.16\ \mathrm{kg \cdot m^2}$<br>$I_y = 4.10\ \mathrm{kg \cdot m^2}, I_z = 4.10\ \mathrm{kg \cdot m^2}$ |
| Added mass | Added-mass coefficients | $X_{udot} = -2.0, Y_{vdot} = -35.0, Z_{wdot} = -35.0$ |

| Notation | Description | Value |
|---|---|---|
| $D(\nu)$ | Hydrodynamic damping coefficients | $K_{pdot} = -0.12, M_{qdot} = -4.5, N_{r\,dot} = -4.5$<br>$X_u = 4.0, Y_v = 8.0, Z_w = 10.0$<br>$X_{\|u\|u} = 18.0, Y_{\|v\|v} = 30.0, Z_{\|w\|w} = 35.0$<br>$K_p = -0.8, M_q = -3.0, N_r = -2.5$ |
| $\nu_{lin}$ | Maximum body-frame velocity components | $u_{max} = 1.5$ m/s, $v_{max} = 0.5$ m/s, $w_{max} = 0.5$ m/s |
| $\nu_{ang}$ | Maximum angular-rate components | $p_{max} = q_{max} = r_{max} = 1.2$ rad/s |
| $\tau$ | Maximum generalized force and moment | $\tau_x = 35.0$ N, $\tau_y = 20.0$ N, $\tau_z = 20.0$ N<br>$\tau_k = 8.0$ N · m, $\tau_m = 10.0$ N · m, $\tau_n = 10.0$ N · m |
| **Initial-condition and target-motion settings** | | |
| Initial separation | Minimum initial inter-AUV separation | at least 0.10 |
| Medium | Medium-stage reset profile | $\|\|p_g(0)\|\| \in [0.25, 0.55]$<br>$\|\|p_i(0) - p_g(0)\|\| \in [0.25, 0.45]$<br>$\|\|v_g(0)\|\| \in [0.003, 0.008]$ |
| Hard | Hard-stage reset profile | $\|\|p_g(0)\|\| \in [0.45, 0.75]$<br>$\|\|p_i(0) - p_g(0)\|\| \in [0.40, 0.70]$<br>$\|\|v_g(0)\|\| \in [0.006, 0.014]$ |
| **Tracking objective, sensing and action interfaces** | | |
| $d^*$ | Desired tracking distance | 0.015 |
| $R_s$ | Effective sensing range | 0.45 |
| Loss threshold | Target-loss distance threshold | 0.65 |
| $o_{i,raw}$ | Raw observation component used in the manuscript | positions, velocities, relative target and neighbor states |
| $o_{i,STG}$ | STG-MAPPO semantic policy input | raw state + tracking diagnostics + semantic graph features |
| $a_i$ | STG-MAPPO velocity-level action interface | $a_i \in [-1, 1]^3$<br>action_scale = 0.75, smoothing = 0.20, $\Delta a_{max} = 0.35$ |
| $a_i$ | Raw baseline low-level action interface | $a_i \in [-1, 1]^6$<br>action_scale = 0.18 |
| **Optimization and training hyperparameters** | | |
| $\gamma$ | Discount factor | 0.95 |
| $\lambda$ | GAE parameter for PPO-family algorithms | 0.95 |
| Actor learning rate | PPO-family actor learning rate | $3.0 \times 10^{-4}$ |
| PPO batch size | Batch size for MAPPO / STG-MAPPO / HAPPO | 800 |
| Q-network batch size | Batch size for MADDPG / MATD3 / MASAC | 1024 |
| Exploration noise | Exploration noise for MADDPG / MATD3 / ATOC | 0.10 / 0.10 / 0.15 |
| Replay buffer | Replay-buffer capacity | $1.0 \times 10^5$ to $2.0 \times 10^5$ |

# B. Medium-Scenario Supplementary Experiments

The main text reports the primary medium-scenario convergence curves and selected final-stage results. Here we provide additional distributional and diagnostic figures that support the interpretation of convergence stability, reward shaping, and seed robustness.

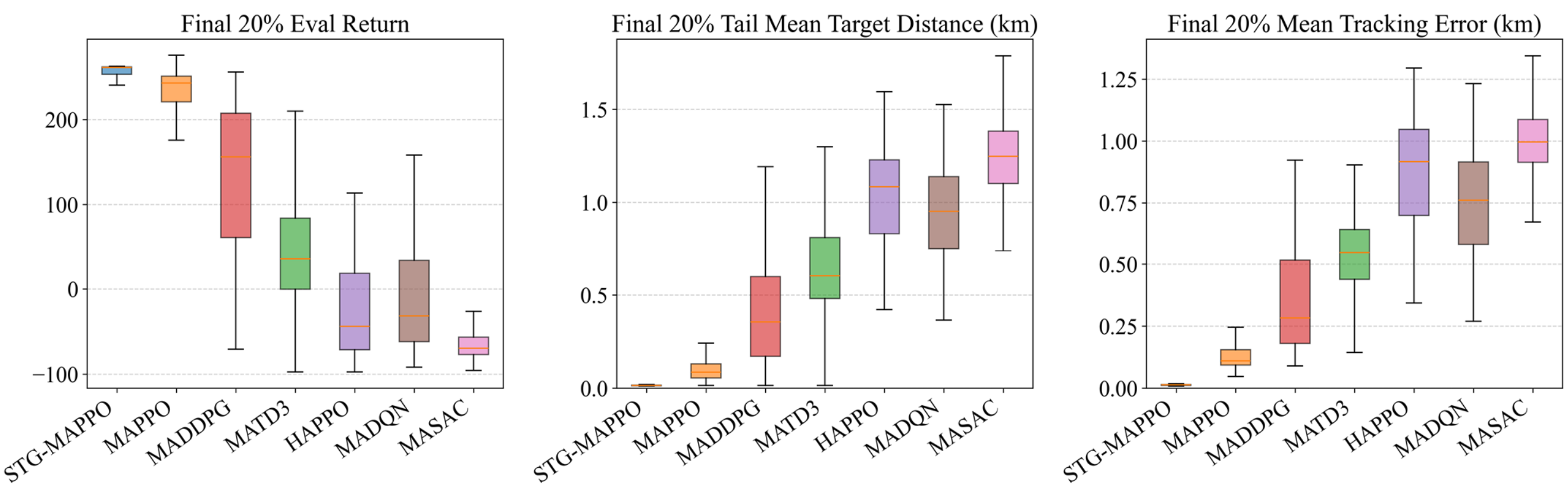

**Figure S1 Medium scenario: final 20% boxplots for main metrics.**

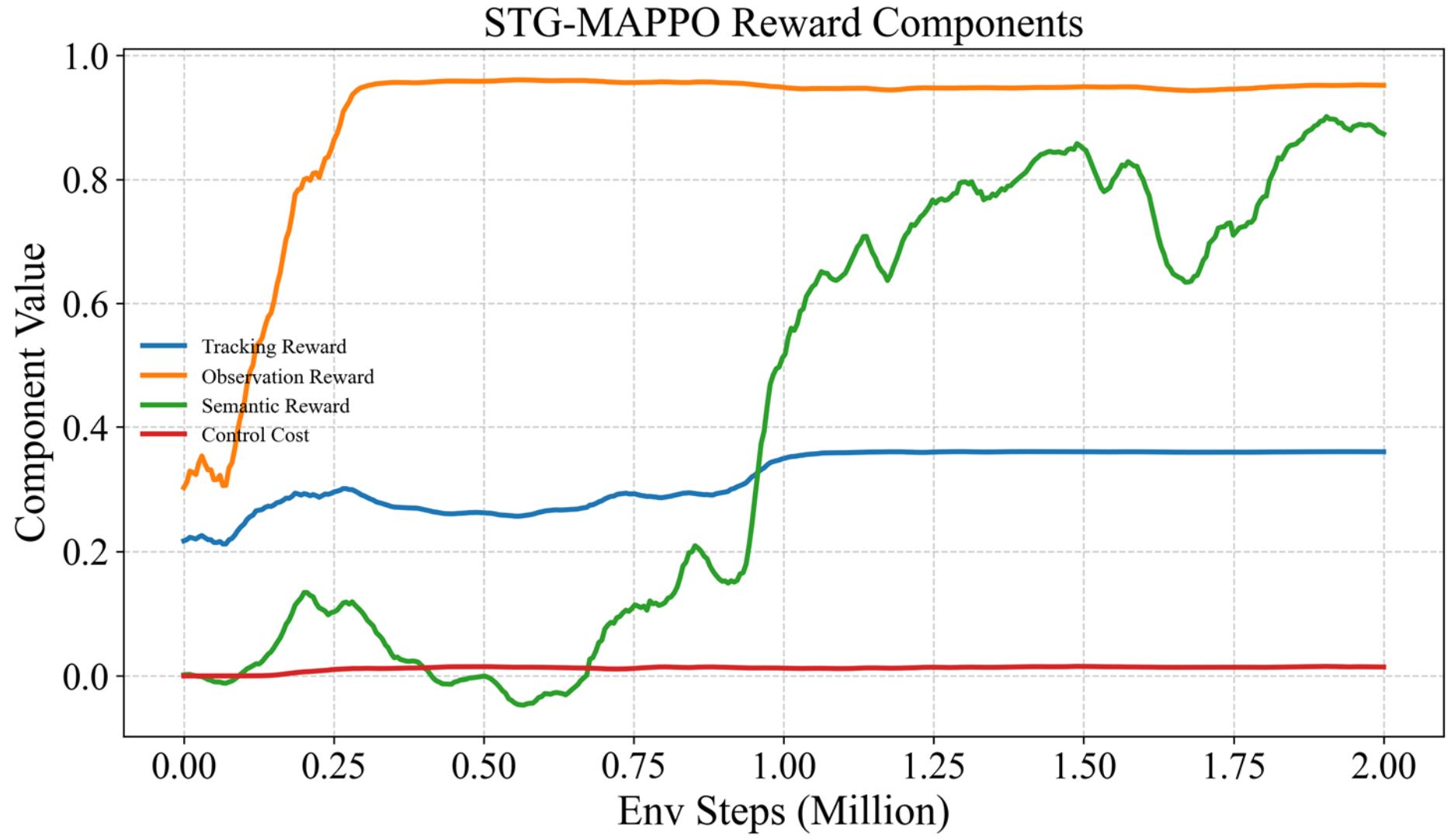

**Figure S2 Medium scenario: STG-MAPPO reward component diagnostics.**

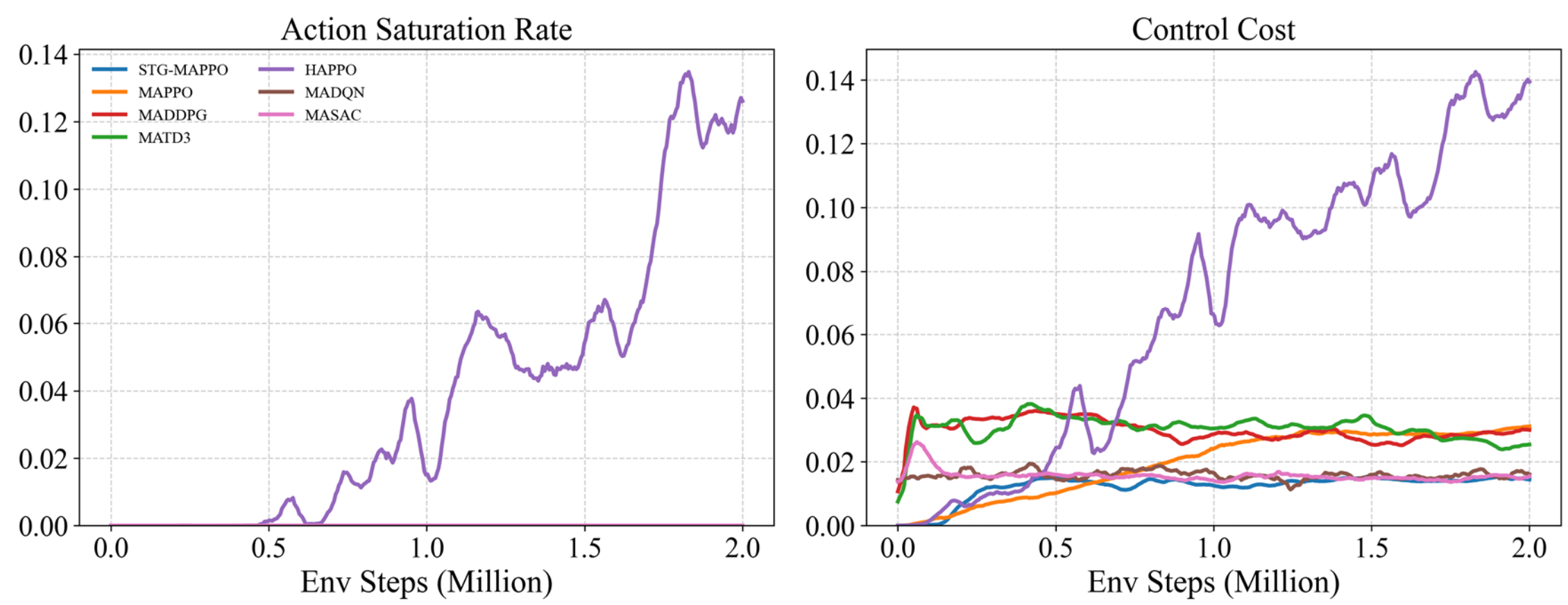

**Figure S3 Medium scenario:** Action saturation and control-cost diagnostics in the medium 4-AUV scenario.

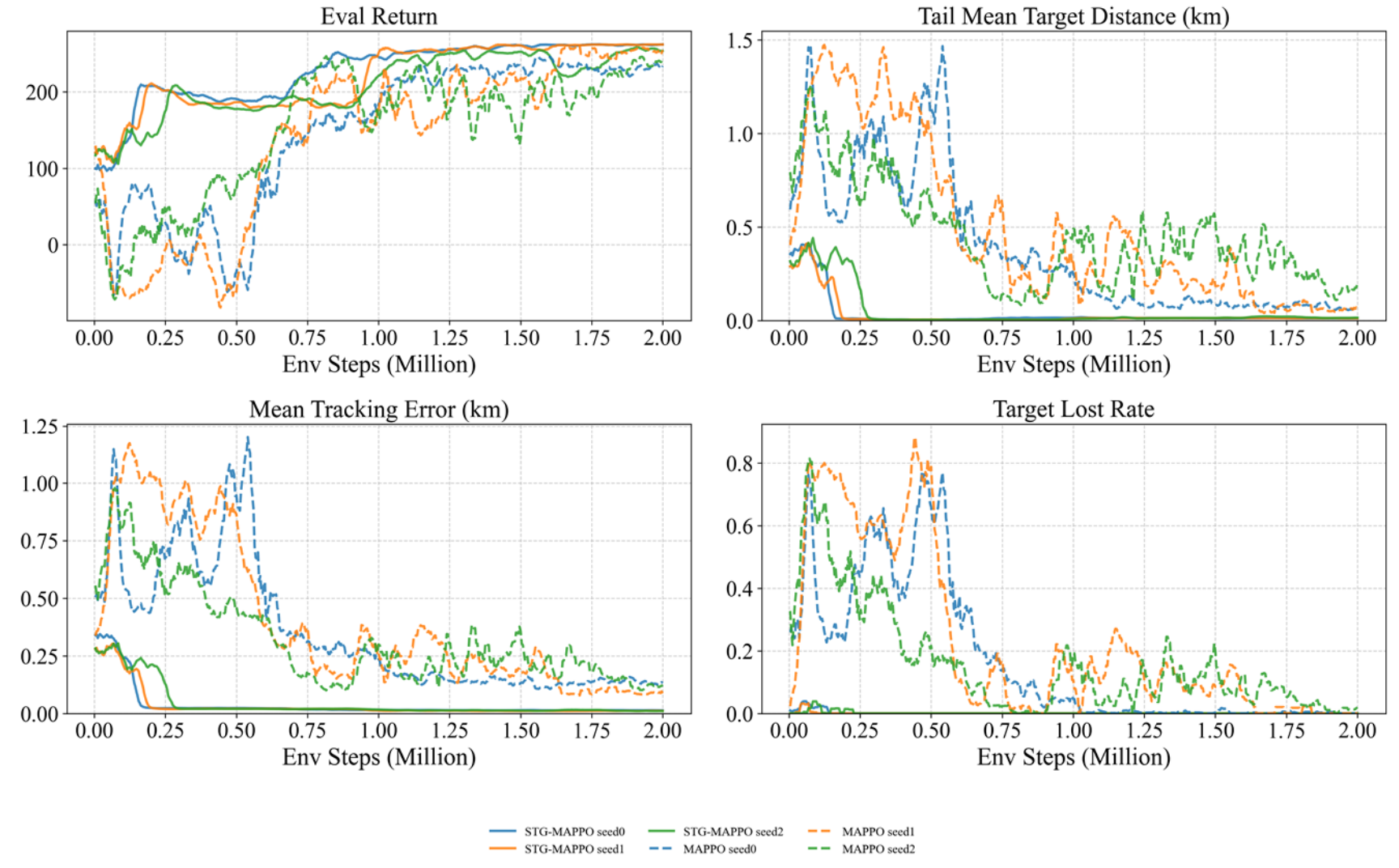

**Figure S4 Medium scenario: per-seed comparison between STG-MAPPO and MAPPO.**

The supplementary medium figures show that the advantage of STG-MAPPO is not only reflected in mean curves. The final-stage distributions remain concentrated near low tracking distance and low target-loss values, and the reward-component plot indicates that semantic shaping contributes without causing action saturation.

**Table S3 Medium scenario: complete final-stage metrics over the last 20% evaluation points.**

| Algorithm | Seeds | Return | Tail dist. | Track err. | Lost | Sat. | Control |
|---|---|---|---|---|---|---|---|
| STG-MAPPO | 3 | 254.81 +/- 11.66 | 0.014 +/- 0.003 | 0.012 +/- 0.001 | 0.000 +/- 0.000 | 0.000 +/- 0.000 | 0.014 +/- 0.006 |
| MAPPO | 3 | 233.06 +/- 17.00 | 0.136 +/- 0.107 | 0.133 +/- 0.035 | 0.018 +/- 0.022 | 0.000 +/- 0.000 | 0.029 +/- 0.004 |
| MADDPG | 3 | 130.89 +/- 91.74 | 0.402 +/- 0.241 | 0.349 +/- 0.207 | 0.130 +/- 0.162 | 0.000 +/- 0.000 | 0.028 +/- 0.001 |
| MATD3 | 3 | 44.69 +/- 49.05 | 0.647 +/- 0.108 | 0.552 +/- 0.130 | 0.367 +/- 0.177 | 0.000 +/- 0.000 | 0.027 +/- 0.004 |
| HAPPO | 3 | -25.42 +/- 62.54 | 1.020 +/- 0.276 | 0.858 +/- 0.249 | 0.640 +/- 0.223 | 0.103 +/- 0.090 | 0.124 +/- 0.062 |
| MADQN | 3 | -13.61 +/- 55.28 | 0.940 +/- 0.194 | 0.747 +/- 0.185 | 0.564 +/- 0.230 | 0.000 +/- 0.000 | 0.016 +/- 0.001 |
| MASAC | 3 | -65.52 +/- 7.39 | 1.239 +/- 0.077 | 0.999 +/- 0.055 | 0.779 +/- 0.037 | 0.000 +/- 0.000 | 0.015 +/- 0.001 |

**Table S4 Medium scenario: per-seed final-stage metrics.**

| Algorithm | Seed | Eval pts | Return | Tail dist. | Track err. | Lost | Sat. | Control |
|---|---|---|---|---|---|---|---|---|
| STG-MAPPO | 0 | 500 | 261.53 | 0.012 | 0.013 | 0.000 | 0.000 | 0.016 |
| STG-MAPPO | 1 | 500 | 261.55 | 0.013 | 0.011 | 0.000 | 0.000 | 0.020 |
| STG-MAPPO | 2 | 500 | 241.35 | 0.018 | 0.013 | 0.000 | 0.000 | 0.008 |
| MAPPO | 0 | 500 | 231.59 | 0.077 | 0.135 | 0.002 | 0.000 | 0.029 |
| MAPPO | 1 | 500 | 250.75 | 0.070 | 0.097 | 0.008 | 0.000 | 0.033 |
| MAPPO | 2 | 500 | 216.84 | 0.259 | 0.168 | 0.043 | 0.000 | 0.026 |
| MADDPG | 0 | 996 | 26.58 | 0.676 | 0.585 | 0.316 | 0.000 | 0.028 |
| MADDPG | 1 | 996 | 167.09 | 0.308 | 0.265 | 0.054 | 0.000 | 0.030 |
| MADDPG | 2 | 996 | 199.01 | 0.222 | 0.197 | 0.019 | 0.000 | 0.028 |
| MATD3 | 0 | 996 | 20.47 | 0.647 | 0.582 | 0.421 | 0.000 | 0.027 |
| MATD3 | 1 | 996 | 12.47 | 0.754 | 0.664 | 0.510 | 0.000 | 0.023 |
| MATD3 | 2 | 996 | 101.15 | 0.539 | 0.410 | 0.169 | 0.000 | 0.031 |
| HAPPO | 0 | 500 | 45.70 | 0.708 | 0.575 | 0.395 | 0.006 | 0.060 |
| HAPPO | 1 | 500 | -71.81 | 1.232 | 1.043 | 0.831 | 0.184 | 0.184 |
| HAPPO | 2 | 500 | -50.16 | 1.120 | 0.957 | 0.693 | 0.119 | 0.129 |
| MADQN | 0 | 500 | -22.54 | 0.929 | 0.749 | 0.582 | 0.000 | 0.016 |
| MADQN | 1 | 500 | -63.88 | 1.140 | 0.930 | 0.785 | 0.000 | 0.015 |
| MADQN | 2 | 500 | 45.59 | 0.752 | 0.560 | 0.327 | 0.000 | 0.016 |
| MASAC | 0 | 996 | -66.72 | 1.178 | 0.970 | 0.779 | 0.000 | 0.013 |
| MASAC | 1 | 996 | -57.60 | 1.213 | 0.964 | 0.741 | 0.000 | 0.015 |
| MASAC | 2 | 996 | -72.24 | 1.326 | 1.062 | 0.816 | 0.000 | 0.016 |

## C. Hard-Scenario Supplementary Experiments

The hard scenario increases target maneuvering difficulty and the initial separation between AUVs and the target. The supplementary hard-scenario figures and tables verify whether each algorithm remains stable under this more demanding setting.

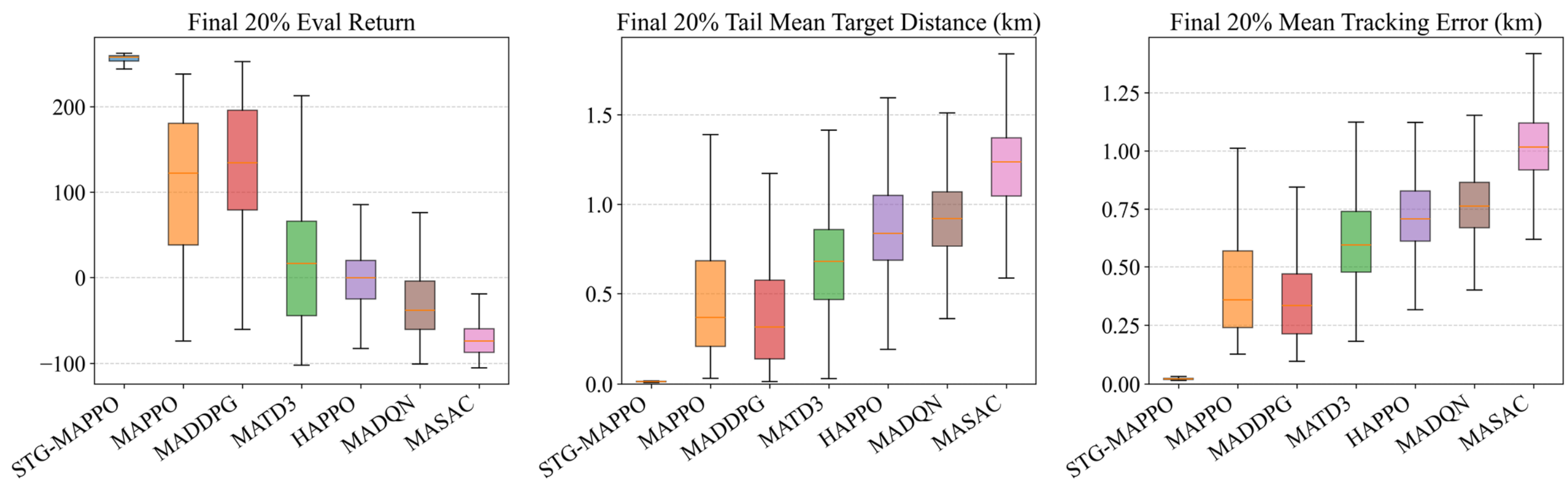

**Figure S5 Hard scenario: final 20% boxplots for main metrics.**

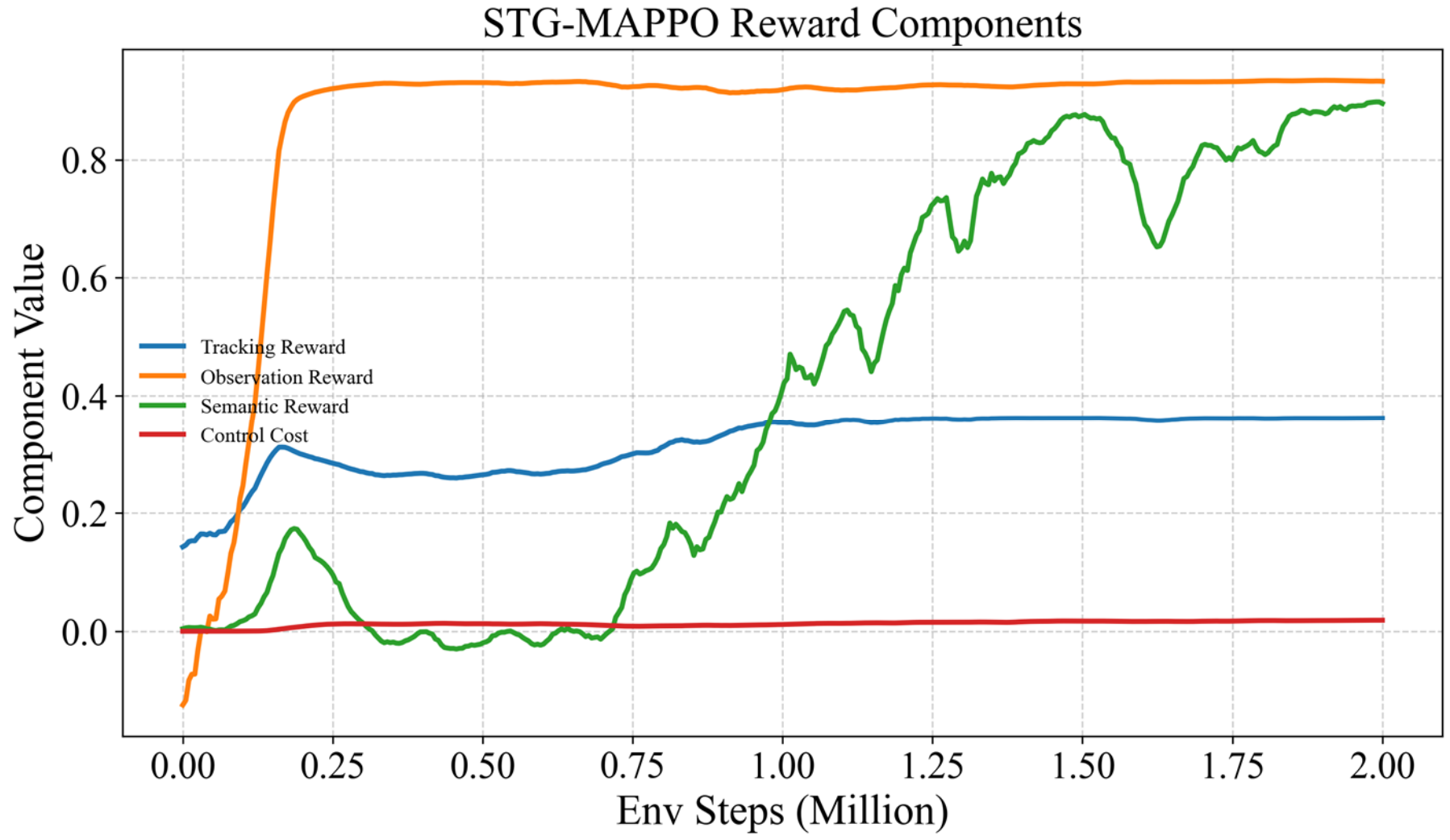

**Figure S6 Hard scenario: STG-MAPPO reward component diagnostics.**

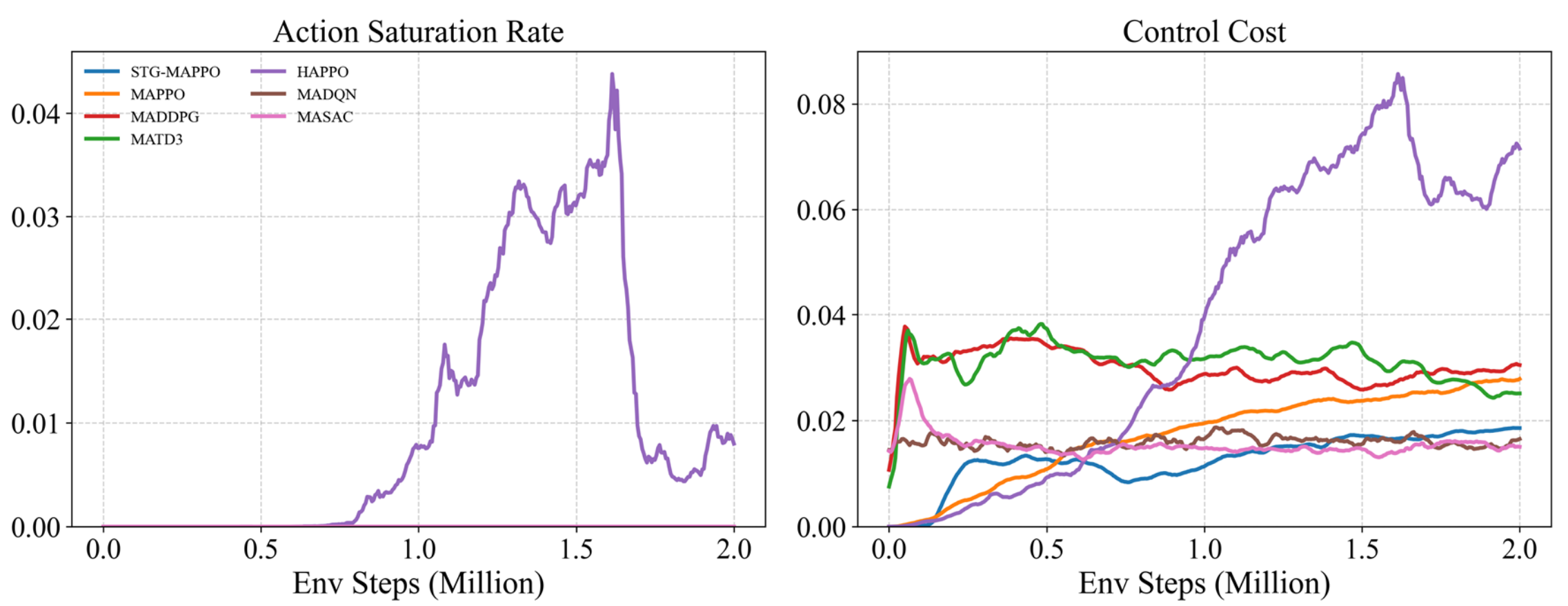

**Figure S7 Hard scenario:** Action saturation and control-cost diagnostics in the hard 4-AUV scenario.

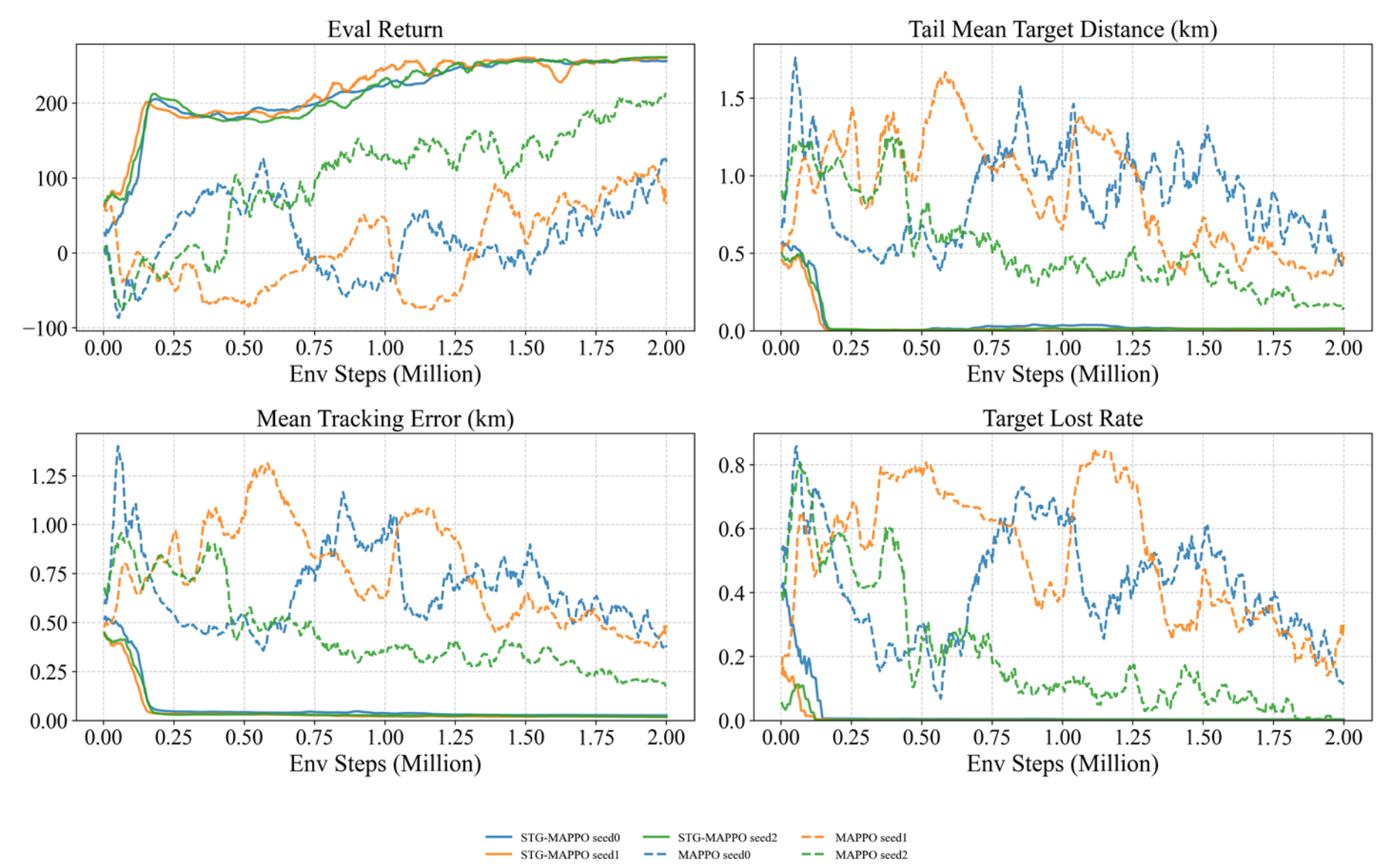

**Figure S8 Hard scenario: per-seed comparison between STG-MAPPO and MAPPO.**

In the hard scenario, several raw baselines show larger tail distances or higher target-loss rates. STG-MAPPO maintains low target distance and stable reward components, indicating that semantic phase information and task-graph diagnostics improve robustness under harder target motion.

**Table S5 Hard scenario: complete final-stage metrics over the last 20% evaluation points.**

| **Algorithm** | **Seeds** | **Return** | **Tail dist.** | **Track err.** | **Lost** | **Sat.** | **Control** |
|---|---|---|---|---|---|---|---|
| STG-MAPPO | 3 | 255.28 +/- 1.90 | 0.012 +/- 0.001 | 0.022 +/- 0.004 | 0.001 +/- 0.001 | 0.000 +/- 0.000 | 0.018 +/- 0.003 |
| MAPPO | 3 | 109.46 +/- 67.36 | 0.452 +/- 0.249 | 0.411 +/- 0.161 | 0.198 +/- 0.147 | 0.000 +/- 0.000 | 0.026 +/- 0.002 |
| MADDPG | 3 | 126.14 +/- 58.74 | 0.374 +/- 0.150 | 0.362 +/- 0.132 | 0.124 +/- 0.099 | 0.000 +/- 0.000 | 0.029 +/- 0.000 |
| MATD3 | 3 | 13.36 +/- 61.51 | 0.695 +/- 0.207 | 0.634 +/- 0.194 | 0.462 +/- 0.270 | 0.000 +/- 0.000 | 0.027 +/- 0.004 |
| HAPPO | 3 | 5.45 +/- 24.47 | 0.854 +/- 0.117 | 0.711 +/- 0.093 | 0.447 +/- 0.081 | 0.011 +/- 0.011 | 0.067 +/- 0.024 |
| MADQN | 3 | -27.22 +/- 22.49 | 0.925 +/- 0.022 | 0.770 +/- 0.052 | 0.596 +/- 0.100 | 0.000 +/- 0.000 | 0.016 +/- 0.002 |
| MASAC | 3 | -70.57 +/- 3.21 | 1.215 +/- 0.078 | 1.029 +/- 0.035 | 0.801 +/- 0.017 | 0.000 +/- 0.000 | 0.015 +/- 0.001 |

**Table S6 Hard scenario: per-seed final-stage metrics.**

| Algorithm | Seed | Eval pts | Return | Tail dist. | Track err. | Lost | Sat. | Control |
|---|---|---|---|---|---|---|---|---|
| STG-MAPPO | 0 | 500 | 255.36 | 0.012 | 0.026 | 0.003 | 0.000 | 0.017 |
| STG-MAPPO | 1 | 500 | 253.34 | 0.011 | 0.020 | 0.001 | 0.000 | 0.021 |
| STG-MAPPO | 2 | 500 | 257.14 | 0.013 | 0.019 | 0.000 | 0.000 | 0.015 |
| MAPPO | 0 | 500 | 58.89 | 0.704 | 0.539 | 0.301 | 0.000 | 0.024 |
| MAPPO | 1 | 500 | 83.57 | 0.448 | 0.463 | 0.263 | 0.000 | 0.028 |
| MAPPO | 2 | 500 | 185.92 | 0.205 | 0.231 | 0.030 | 0.000 | 0.027 |
| MADDPG | 0 | 996 | 73.69 | 0.498 | 0.486 | 0.226 | 0.000 | 0.029 |
| MADDPG | 1 | 996 | 115.12 | 0.416 | 0.375 | 0.118 | 0.000 | 0.030 |
| MADDPG | 2 | 996 | 189.62 | 0.208 | 0.224 | 0.028 | 0.000 | 0.029 |
| MATD3 | 0 | 996 | 8.62 | 0.607 | 0.591 | 0.466 | 0.000 | 0.028 |
| MATD3 | 1 | 996 | -45.64 | 0.932 | 0.845 | 0.730 | 0.000 | 0.023 |
| MATD3 | 2 | 996 | 77.11 | 0.547 | 0.465 | 0.190 | 0.000 | 0.032 |
| HAPPO | 0 | 500 | 33.39 | 0.739 | 0.608 | 0.354 | 0.003 | 0.047 |
| HAPPO | 1 | 500 | -4.90 | 0.849 | 0.737 | 0.504 | 0.008 | 0.061 |
| HAPPO | 2 | 500 | -12.15 | 0.973 | 0.789 | 0.484 | 0.023 | 0.093 |
| MADQN | 0 | 500 | -7.75 | 0.901 | 0.711 | 0.518 | 0.000 | 0.018 |
| MADQN | 1 | 500 | -51.83 | 0.931 | 0.807 | 0.709 | 0.000 | 0.015 |
| MADQN | 2 | 500 | -22.07 | 0.944 | 0.791 | 0.562 | 0.000 | 0.014 |
| MASAC | 0 | 996 | -67.19 | 1.181 | 0.994 | 0.782 | 0.000 | 0.015 |
| MASAC | 1 | 996 | -70.93 | 1.160 | 1.028 | 0.808 | 0.000 | 0.016 |
| MASAC | 2 | 996 | -73.59 | 1.304 | 1.065 | 0.813 | 0.000 | 0.015 |

# D. Best-Checkpoint Stress-Test Supplement

The stress-test results reload the best checkpoint of each algorithm and evaluate it under multiple perturbations. Unlike training-curve statistics, this procedure directly tests whether a saved policy generalizes to modified target speed, initialization range, sensing range, and communication quality.

**Table S7 Medium-trained best checkpoints under all medium stress conditions.**

| Algorithm | Condition | Return | Tail dist. | Track err. | Lost | Action | Sat. | Control | Comm. |
|---|---|---|---|---|---|---|---|---|---|
| STG-MAPPO | Nominal-medium | 1039.2 +/- 5.0 | 0.013 +/- 0.001 | 0.003 +/- 0.001 | 0.000 +/- 0.000 | 0.250 +/- 0.020 | 0.000 +/- 0.000 | 0.015 +/- 0.003 | 0.979 +/- 0.005 |
| STG-MAPPO | Fast target | 1017.1 +/- 34.0 | 0.014 +/- 0.003 | 0.003 +/- 0.002 | 0.000 +/- 0.000 | 0.241 +/- 0.031 | 0.000 +/- 0.000 | 0.015 +/- 0.003 | 0.978 +/- 0.005 |
| STG-MAPPO | Far initialization | 1011.2 +/- 35.0 | 0.013 +/- 0.003 | 0.003 +/- 0.002 | 0.000 +/- 0.000 | 0.247 +/- 0.018 | 0.000 +/- 0.000 | 0.015 +/- 0.003 | 0.957 +/- 0.010 |
| STG-MAPPO | Limited sensing | 1031.4 +/- 4.9 | 0.013 +/- 0.001 | 0.003 +/- 0.001 | 0.000 +/- 0.000 | 0.253 +/- 0.020 | 0.000 +/- 0.000 | 0.015 +/- 0.003 | 0.969 +/- 0.007 |
| STG-MAPPO | Communication degraded | 1012.1 +/- 4.8 | 0.013 +/- 0.001 | 0.003 +/- 0.001 | 0.000 +/- 0.000 | 0.253 +/- 0.020 | 0.000 +/- 0.000 | 0.015 +/- 0.003 | 0.969 +/- 0.007 |
| STG-MAPPO | Combined stress | 1002.0 +/- 44.1 | 0.013 +/- 0.003 | 0.003 +/- 0.002 | 0.000 +/- 0.000 | 0.243 +/- 0.027 | 0.000 +/- 0.000 | 0.015 +/- 0.003 | 0.945 +/- 0.014 |
| MAPPO | Nominal-medium | 798.4 +/- 362.8 | 0.185 +/- 0.198 | 0.171 +/- 0.197 | 0.052 +/- 0.157 | 0.297 +/- 0.032 | 0.000 +/- 0.000 | 0.010 +/- 0.002 | 0.610 +/- 0.306 |
| MAPPO | Fast target | 767.4 +/- 430.6 | 0.205 +/- 0.253 | 0.191 +/- 0.253 | 0.111 +/- 0.333 | 0.298 +/- 0.029 | 0.000 +/- 0.000 | 0.010 +/- 0.002 | 0.596 +/- 0.328 |
| MAPPO | Far initialization | 697.1 +/- 393.2 | 0.215 +/- 0.235 | 0.200 +/- 0.235 | 0.107 +/- 0.287 | 0.297 +/- 0.032 | 0.000 +/- 0.000 | 0.010 +/- 0.002 | 0.509 +/- 0.290 |
| MAPPO | Limited sensing | 741.0 +/- 358.5 | 0.185 +/- 0.198 | 0.171 +/- 0.197 | 0.052 +/- 0.157 | 0.297 +/- 0.032 | 0.000 +/- 0.000 | 0.010 +/- 0.002 | 0.490 +/- 0.330 |
| MAPPO | Communication degraded | 692.1 +/- 328.0 | 0.185 +/- 0.198 | 0.171 +/- 0.197 | 0.052 +/- 0.157 | 0.297 +/- 0.032 | 0.000 +/- 0.000 | 0.010 +/- 0.002 | 0.490 +/- 0.330 |
| MAPPO | Combined stress | 627.1 +/- 391.2 | 0.221 +/- 0.244 | 0.206 +/- 0.244 | 0.121 +/- 0.301 | 0.300 +/- 0.029 | 0.000 +/- 0.000 | 0.011 +/- 0.002 | 0.380 +/- 0.300 |
| MADDPG | Nominal-medium | 854.8 +/- 324.8 | 0.175 +/- 0.237 | 0.166 +/- 0.233 | 0.048 +/- 0.144 | 0.334 +/- 0.055 | 0.000 +/- 0.000 | 0.013 +/- 0.005 | 0.644 +/- 0.330 |
| MADDPG | Fast target | 818.9 +/- 363.8 | 0.211 +/- 0.286 | 0.200 +/- 0.282 | 0.107 +/- 0.241 | 0.334 +/- 0.052 | 0.000 +/- 0.000 | 0.013 +/- 0.004 | 0.621 +/- 0.344 |
| MADDPG | Far initialization | 676.7 +/- 483.3 | 0.251 +/- 0.347 | 0.240 +/- 0.344 | 0.174 +/- 0.333 | 0.336 +/- 0.047 | 0.000 +/- 0.000 | 0.013 +/- 0.004 | 0.513 +/- 0.375 |
| MADDPG | Limited sensing | 802.0 +/- 338.0 | 0.175 +/- 0.237 | 0.166 +/- 0.233 | 0.048 +/- 0.144 | 0.334 +/- 0.055 | 0.000 +/- 0.000 | 0.013 +/- 0.005 | 0.532 +/- 0.383 |
| MADDPG | Communication degraded | 748.8 +/- 300.6 | 0.175 +/- 0.237 | 0.166 +/- 0.233 | 0.048 +/- 0.144 | 0.334 +/- 0.055 | 0.000 +/- 0.000 | 0.013 +/- 0.005 | 0.532 +/- 0.383 |
| MADDPG | Combined stress | 637.1 +/- 469.0 | 0.256 +/- 0.349 | 0.244 +/- 0.347 | 0.178 +/- 0.335 | 0.337 +/- 0.048 | 0.000 +/- 0.000 | 0.013 +/- 0.004 | 0.443 +/- 0.337 |
| MATD3 | Nominal-medium | 343.3 +/- 330.0 | 0.594 +/- 0.290 | 0.581 +/- 0.290 | 0.440 +/- 0.287 | 0.388 +/- 0.026 | 0.000 +/- 0.000 | 0.016 +/- 0.002 | 0.164 +/- 0.163 |
| MATD3 | Fast target | 323.0 +/- 336.9 | 0.612 +/- 0.323 | 0.597 +/- 0.322 | 0.408 +/- 0.251 | 0.387 +/- 0.021 | 0.000 +/- 0.000 | 0.016 +/- 0.002 | 0.163 +/- 0.188 |
| MATD3 | Far initialization | -18.5 +/- 273.1 | 0.972 +/- 0.378 | 0.957 +/- 0.378 | 0.643 +/- 0.319 | 0.403 +/- 0.021 | 0.000 +/- 0.000 | 0.017 +/- 0.002 | 0.014 +/- 0.037 |
| MATD3 | Limited sensing | 297.8 +/- 307.6 | 0.594 +/- 0.290 | 0.581 +/- 0.290 | 0.440 +/- 0.287 | 0.388 +/- 0.026 | 0.000 +/- 0.000 | 0.016 +/- 0.002 | 0.042 +/- 0.084 |
| MATD3 | Communication degraded | 293.5 +/- 302.9 | 0.594 +/- 0.290 | 0.581 +/- 0.290 | 0.440 +/- 0.287 | 0.388 +/- 0.026 | 0.000 +/- 0.000 | 0.016 +/- 0.002 | 0.042 +/- 0.084 |
| MATD3 | Combined stress | -35.5 +/- 259.0 | 0.965 +/- 0.382 | 0.950 +/- 0.381 | 0.646 +/- 0.324 | 0.404 +/- 0.018 | 0.000 +/- 0.000 | 0.017 +/- 0.002 | 0.000 +/- 0.000 |
| HAPPO | Nominal-medium | 472.9 +/- 83.1 | 0.421 +/- 0.066 | 0.406 +/- 0.066 | 0.019 +/- 0.037 | 0.000 +/- 0.000 | 0.000 +/- 0.000 | 0.000 +/- 0.000 | 0.204 +/- 0.082 |
| HAPPO | Fast target | 303.6 +/- 124.5 | 0.598 +/- 0.093 | 0.583 +/- 0.093 | 0.411 +/- 0.227 | 0.000 +/- 0.000 | 0.000 +/- 0.000 | 0.000 +/- 0.000 | 0.115 +/- 0.069 |
| HAPPO | Far initialization | 44.6 +/- 145.2 | 0.685 +/- 0.088 | 0.670 +/- 0.088 | 0.549 +/- 0.229 | 0.000 +/- 0.000 | 0.000 +/- 0.000 | 0.000 +/- 0.000 | 0.000 +/- 0.000 |
| HAPPO | Limited sensing | 387.8 +/- 54.6 | 0.421 +/- 0.066 | 0.406 +/- 0.066 | 0.019 +/- 0.037 | 0.000 +/- 0.000 | 0.000 +/- 0.000 | 0.000 +/- 0.000 | 0.002 +/- 0.007 |
| HAPPO | Communication degraded | 387.6 +/- 54.2 | 0.421 +/- 0.066 | 0.406 +/- 0.066 | 0.019 +/- 0.037 | 0.000 +/- 0.000 | 0.000 +/- 0.000 | 0.000 +/- 0.000 | 0.002 +/- 0.007 |
| HAPPO | Combined stress | 11.3 +/- 135.8 | 0.713 +/- 0.085 | 0.698 +/- 0.085 | 0.617 +/- 0.208 | 0.000 +/- 0.000 | 0.000 +/- 0.000 | 0.000 +/- 0.000 | 0.000 +/- 0.000 |
| MADQN | Nominal-medium | 142.5 +/- 152.2 | 0.680 +/- 0.144 | 0.665 +/- 0.144 | 0.593 +/- 0.296 | 0.054 +/- 0.005 | 0.000 +/- 0.000 | 0.000 +/- 0.000 | 0.074 +/- 0.050 |
| MADQN | Fast target | 118.9 +/- 187.4 | 0.713 +/- 0.193 | 0.698 +/- 0.193 | 0.581 +/- 0.301 | 0.055 +/- 0.008 | 0.000 +/- 0.000 | 0.000 +/- 0.000 | 0.069 +/- 0.049 |
| MADQN | Far initialization | 13.7 +/- 167.0 | 0.733 +/- 0.185 | 0.718 +/- 0.185 | 0.630 +/- 0.235 | 0.050 +/- 0.010 | 0.000 +/- 0.000 | 0.000 +/- 0.000 | 0.013 +/- 0.021 |
| MADQN | Limited sensing | 109.6 +/- 142.0 | 0.680 +/- 0.144 | 0.665 +/- 0.144 | 0.593 +/- 0.296 | 0.054 +/- 0.005 | 0.000 +/- 0.000 | 0.000 +/- 0.000 | 0.013 +/- 0.026 |
| MADQN | Communication degraded | 108.3 +/- 142.2 | 0.680 +/- 0.144 | 0.665 +/- 0.144 | 0.593 +/- 0.296 | 0.054 +/- 0.005 | 0.000 +/- 0.000 | 0.000 +/- 0.000 | 0.013 +/- 0.026 |
| MADQN | Combined stress | -14.1 +/- 155.4 | 0.749 +/- 0.182 | 0.734 +/- 0.182 | 0.611 +/- 0.234 | 0.050 +/- 0.009 | 0.000 +/- 0.000 | 0.000 +/- 0.000 | 0.000 +/- 0.001 |
| MASAC | Nominal-medium | -241.9 +/- 73.9 | 1.172 +/- 0.090 | 1.157 +/- 0.090 | 0.938 +/- 0.108 | 0.064 +/- 0.010 | 0.000 +/- 0.000 | 0.000 +/- 0.000 | 0.026 +/- 0.013 |
| MASAC | Fast target | -226.4 +/- 74.8 | 1.195 +/- 0.125 | 1.180 +/- 0.125 | 0.873 +/- 0.173 | 0.064 +/- 0.009 | 0.000 +/- 0.000 | 0.000 +/- 0.000 | 0.026 +/- 0.013 |

| MASAC | Far initialization | -278.6 +/- 93.8 | 1.205 +/- 0.152 | 1.190 +/- 0.152 | 0.889 +/- 0.132 | 0.063 +/- 0.006 | 0.000 +/- 0.000 | 0.000 +/- 0.000 | 0.000 +/- 0.000 |
|---|---|---|---|---|---|---|---|---|---|
| MASAC | Limited sensing | -255.9 +/- 70.7 | 1.172 +/- 0.090 | 1.157 +/- 0.090 | 0.938 +/- 0.108 | 0.064 +/- 0.010 | 0.000 +/- 0.000 | 0.000 +/- 0.000 | 0.000 +/- 0.001 |
| MASAC | Communication degraded | -255.9 +/- 70.7 | 1.172 +/- 0.090 | 1.157 +/- 0.090 | 0.938 +/- 0.108 | 0.064 +/- 0.010 | 0.000 +/- 0.000 | 0.000 +/- 0.000 | 0.000 +/- 0.001 |
| MASAC | Combined stress | -277.9 +/- 91.2 | 1.219 +/- 0.164 | 1.204 +/- 0.164 | 0.875 +/- 0.125 | 0.063 +/- 0.007 | 0.000 +/- 0.000 | 0.000 +/- 0.000 | 0.000 +/- 0.000 |

**Table S8 Hard-trained best checkpoints under all hard stress conditions.**

| Algorithm | Condition | Return | Tail dist. | Track err. | Lost | Action | Sat. | Control | Comm. |
|---|---|---|---|---|---|---|---|---|---|
| STG-MAPPO | Nominal-hard | 994.2 +/- 68.9 | 0.013 +/- 0.003 | 0.003 +/- 0.002 | 0.000 +/- 0.000 | 0.241 +/- 0.050 | 0.000 +/- 0.000 | 0.015 +/- 0.005 | 0.956 +/- 0.013 |
| STG-MAPPO | Faster target | 979.9 +/- 78.0 | 0.014 +/- 0.006 | 0.005 +/- 0.004 | 0.000 +/- 0.000 | 0.233 +/- 0.053 | 0.000 +/- 0.000 | 0.015 +/- 0.005 | 0.954 +/- 0.012 |
| STG-MAPPO | Farther initialization | 994.2 +/- 68.9 | 0.013 +/- 0.003 | 0.003 +/- 0.002 | 0.000 +/- 0.000 | 0.241 +/- 0.050 | 0.000 +/- 0.000 | 0.015 +/- 0.005 | 0.956 +/- 0.013 |
| STG-MAPPO | Limited sensing | 984.6 +/- 71.4 | 0.012 +/- 0.003 | 0.003 +/- 0.002 | 0.000 +/- 0.000 | 0.245 +/- 0.046 | 0.000 +/- 0.000 | 0.016 +/- 0.005 | 0.942 +/- 0.018 |
| STG-MAPPO | Communication degraded | 965.8 +/- 71.1 | 0.012 +/- 0.003 | 0.003 +/- 0.002 | 0.000 +/- 0.000 | 0.245 +/- 0.046 | 0.000 +/- 0.000 | 0.016 +/- 0.005 | 0.942 +/- 0.018 |
| STG-MAPPO | Combined stress | 971.5 +/- 83.2 | 0.014 +/- 0.006 | 0.005 +/- 0.004 | 0.000 +/- 0.000 | 0.237 +/- 0.049 | 0.000 +/- 0.000 | 0.015 +/- 0.005 | 0.941 +/- 0.017 |
| MAPPO | Nominal-hard | 950.8 +/- 97.6 | 0.064 +/- 0.032 | 0.050 +/- 0.031 | 0.000 +/- 0.000 | 0.314 +/- 0.019 | 0.000 +/- 0.000 | 0.012 +/- 0.001 | 0.724 +/- 0.110 |
| MAPPO | Faster target | 911.1 +/- 137.5 | 0.095 +/- 0.088 | 0.081 +/- 0.087 | 0.002 +/- 0.006 | 0.312 +/- 0.020 | 0.000 +/- 0.000 | 0.012 +/- 0.001 | 0.682 +/- 0.151 |
| MAPPO | Farther initialization | 950.8 +/- 97.6 | 0.064 +/- 0.032 | 0.050 +/- 0.031 | 0.000 +/- 0.000 | 0.314 +/- 0.019 | 0.000 +/- 0.000 | 0.012 +/- 0.001 | 0.724 +/- 0.110 |
| MAPPO | Limited sensing | 898.6 +/- 114.7 | 0.064 +/- 0.032 | 0.050 +/- 0.031 | 0.000 +/- 0.000 | 0.314 +/- 0.019 | 0.000 +/- 0.000 | 0.012 +/- 0.001 | 0.619 +/- 0.147 |
| MAPPO | Communication degraded | 836.7 +/- 100.0 | 0.064 +/- 0.032 | 0.050 +/- 0.031 | 0.000 +/- 0.000 | 0.314 +/- 0.019 | 0.000 +/- 0.000 | 0.012 +/- 0.001 | 0.619 +/- 0.147 |
| MAPPO | Combined stress | 850.8 +/- 161.5 | 0.095 +/- 0.088 | 0.081 +/- 0.087 | 0.002 +/- 0.006 | 0.312 +/- 0.020 | 0.000 +/- 0.000 | 0.012 +/- 0.001 | 0.557 +/- 0.208 |
| MADDPG | Nominal-hard | 645.8 +/- 359.3 | 0.346 +/- 0.365 | 0.331 +/- 0.365 | 0.187 +/- 0.305 | 0.343 +/- 0.048 | 0.000 +/- 0.000 | 0.014 +/- 0.003 | 0.454 +/- 0.301 |
| MADDPG | Faster target | 649.6 +/- 390.4 | 0.282 +/- 0.407 | 0.268 +/- 0.407 | 0.139 +/- 0.277 | 0.349 +/- 0.043 | 0.000 +/- 0.000 | 0.014 +/- 0.003 | 0.466 +/- 0.283 |
| MADDPG | Farther initialization | 645.8 +/- 359.3 | 0.346 +/- 0.365 | 0.331 +/- 0.365 | 0.187 +/- 0.305 | 0.343 +/- 0.048 | 0.000 +/- 0.000 | 0.014 +/- 0.003 | 0.454 +/- 0.301 |
| MADDPG | Limited sensing | 589.3 +/- 353.4 | 0.346 +/- 0.365 | 0.331 +/- 0.365 | 0.187 +/- 0.305 | 0.343 +/- 0.048 | 0.000 +/- 0.000 | 0.014 +/- 0.003 | 0.340 +/- 0.291 |
| MADDPG | Communication degraded | 555.3 +/- 326.4 | 0.346 +/- 0.365 | 0.331 +/- 0.365 | 0.187 +/- 0.305 | 0.343 +/- 0.048 | 0.000 +/- 0.000 | 0.014 +/- 0.003 | 0.340 +/- 0.291 |
| MADDPG | Combined stress | 594.2 +/- 373.7 | 0.282 +/- 0.407 | 0.268 +/- 0.407 | 0.139 +/- 0.277 | 0.349 +/- 0.043 | 0.000 +/- 0.000 | 0.014 +/- 0.003 | 0.350 +/- 0.245 |
| MATD3 | Nominal-hard | 301.0 +/- 407.0 | 0.554 +/- 0.449 | 0.539 +/- 0.448 | 0.332 +/- 0.392 | 0.391 +/- 0.025 | 0.000 +/- 0.000 | 0.017 +/- 0.002 | 0.205 +/- 0.244 |
| MATD3 | Faster target | 295.5 +/- 380.8 | 0.584 +/- 0.452 | 0.569 +/- 0.452 | 0.308 +/- 0.414 | 0.393 +/- 0.029 | 0.000 +/- 0.000 | 0.017 +/- 0.002 | 0.188 +/- 0.200 |
| MATD3 | Farther initialization | 301.0 +/- 407.0 | 0.554 +/- 0.449 | 0.539 +/- 0.448 | 0.332 +/- 0.392 | 0.391 +/- 0.025 | 0.000 +/- 0.000 | 0.017 +/- 0.002 | 0.205 +/- 0.244 |
| MATD3 | Limited sensing | 260.3 +/- 379.0 | 0.554 +/- 0.449 | 0.539 +/- 0.448 | 0.332 +/- 0.392 | 0.391 +/- 0.025 | 0.000 +/- 0.000 | 0.017 +/- 0.002 | 0.131 +/- 0.173 |
| MATD3 | Communication degraded | 247.3 +/- 363.7 | 0.554 +/- 0.449 | 0.539 +/- 0.448 | 0.332 +/- 0.392 | 0.391 +/- 0.025 | 0.000 +/- 0.000 | 0.017 +/- 0.002 | 0.131 +/- 0.173 |
| MATD3 | Combined stress | 245.0 +/- 350.7 | 0.584 +/- 0.452 | 0.569 +/- 0.452 | 0.308 +/- 0.414 | 0.393 +/- 0.029 | 0.000 +/- 0.000 | 0.017 +/- 0.002 | 0.090 +/- 0.114 |
| HAPPO | Nominal-hard | 271.9 +/- 294.0 | 0.518 +/- 0.237 | 0.503 +/- 0.237 | 0.361 +/- 0.269 | 0.074 +/- 0.111 | 0.000 +/- 0.000 | 0.002 +/- 0.003 | 0.128 +/- 0.202 |
| HAPPO | Faster target | 161.8 +/- 353.5 | 0.674 +/- 0.345 | 0.659 +/- 0.345 | 0.576 +/- 0.392 | 0.075 +/- 0.113 | 0.000 +/- 0.000 | 0.002 +/- 0.003 | 0.118 +/- 0.198 |
| HAPPO | Farther initialization | 271.9 +/- 294.0 | 0.518 +/- 0.237 | 0.503 +/- 0.237 | 0.361 +/- 0.269 | 0.074 +/- 0.111 | 0.000 +/- 0.000 | 0.002 +/- 0.003 | 0.128 +/- 0.202 |
| HAPPO | Limited sensing | 240.6 +/- 266.0 | 0.518 +/- 0.237 | 0.503 +/- 0.237 | 0.361 +/- 0.269 | 0.074 +/- 0.111 | 0.000 +/- 0.000 | 0.002 +/- 0.003 | 0.060 +/- 0.120 |
| HAPPO | Communication degraded | 234.6 +/- 256.2 | 0.518 +/- 0.237 | 0.503 +/- 0.237 | 0.361 +/- 0.269 | 0.074 +/- 0.111 | 0.000 +/- 0.000 | 0.002 +/- 0.003 | 0.060 +/- 0.120 |
| HAPPO | Combined stress | 131.3 +/- 323.0 | 0.674 +/- 0.345 | 0.659 +/- 0.345 | 0.576 +/- 0.392 | 0.075 +/- 0.113 | 0.000 +/- 0.000 | 0.002 +/- 0.003 | 0.061 +/- 0.122 |
| MADQN | Nominal-hard | 45.9 +/- 241.0 | 0.735 +/- 0.337 | 0.720 +/- 0.337 | 0.514 +/- 0.343 | 0.056 +/- 0.005 | 0.000 +/- 0.000 | 0.001 +/- 0.000 | 0.011 +/- 0.026 |
| MADQN | Faster target | 48.3 +/- 256.9 | 0.812 +/- 0.397 | 0.797 +/- 0.397 | 0.499 +/- 0.345 | 0.051 +/- 0.009 | 0.000 +/- 0.000 | 0.000 +/- 0.000 | 0.019 +/- 0.044 |
| MADQN | Farther initialization | 45.9 +/- 241.0 | 0.735 +/- 0.337 | 0.720 +/- 0.337 | 0.514 +/- 0.343 | 0.056 +/- 0.005 | 0.000 +/- 0.000 | 0.001 +/- 0.000 | 0.011 +/- 0.026 |
| MADQN | Limited sensing | 27.0 +/- 228.8 | 0.735 +/- 0.337 | 0.720 +/- 0.337 | 0.514 +/- 0.343 | 0.056 +/- 0.005 | 0.000 +/- 0.000 | 0.001 +/- 0.000 | 0.000 +/- 0.000 |
| MADQN | Communication degraded | 27.0 +/- 228.8 | 0.735 +/- 0.337 | 0.720 +/- 0.337 | 0.514 +/- 0.343 | 0.056 +/- 0.005 | 0.000 +/- 0.000 | 0.001 +/- 0.000 | 0.000 +/- 0.000 |
| MADQN | Combined stress | 25.0 +/- 241.0 | 0.812 +/- 0.397 | 0.797 +/- 0.397 | 0.499 +/- 0.345 | 0.051 +/- 0.009 | 0.000 +/- 0.000 | 0.000 +/- 0.000 | 0.001 +/- 0.002 |
| MASAC | Nominal-hard | -275.2 +/- 86.2 | 1.200 +/- 0.146 | 1.185 +/- 0.146 | 0.839 +/- 0.165 | 0.062 +/- 0.007 | 0.000 +/- 0.000 | 0.000 +/- 0.000 | 0.000 +/- 0.001 |
| MASAC | Faster target | -268.5 +/- 79.7 | 1.254 +/- 0.164 | 1.239 +/- 0.164 | 0.881 +/- 0.126 | 0.062 +/- 0.006 | 0.000 +/- 0.000 | 0.000 +/- 0.000 | 0.000 +/- 0.001 |
| MASAC | Farther initialization | -275.2 +/- 86.2 | 1.200 +/- 0.146 | 1.185 +/- 0.146 | 0.839 +/- 0.165 | 0.062 +/- 0.007 | 0.000 +/- 0.000 | 0.000 +/- 0.000 | 0.000 +/- 0.001 |
| MASAC | Limited sensing | -277.7 +/- 84.3 | 1.200 +/- 0.146 | 1.185 +/- 0.146 | 0.839 +/- 0.165 | 0.062 +/- 0.007 | 0.000 +/- 0.000 | 0.000 +/- 0.000 | 0.000 +/- 0.000 |
| MASAC | Communication degraded | -277.7 +/- 84.3 | 1.200 +/- 0.146 | 1.185 +/- 0.146 | 0.839 +/- 0.165 | 0.062 +/- 0.007 | 0.000 +/- 0.000 | 0.000 +/- 0.000 | 0.000 +/- 0.000 |
| MASAC | Combined stress | -274.4 +/- 74.6 | 1.254 +/- 0.164 | 1.239 +/- 0.164 | 0.881 +/- 0.126 | 0.062 +/- 0.006 | 0.000 +/- 0.000 | 0.000 +/- 0.000 | 0.000 +/- 0.000 |

The full stress-test tables show that STG-MAPPO preserves low tail distance and target-loss rate across all perturbation types. This is important because the proposed method is intended for autonomous AUV networking rather than a single fixed training distribution.

# E. Ablation Supplementary Experiments

The ablation study decomposes the proposed method into action abstraction and semantic information. MAPPO-velocity3-nonsemantic tests whether the low-dimensional velocity action alone explains the improvement. MAPPO-semantic-state-only tests semantic observations without semantic reward. STG-MAPPO-full combines semantic state, semantic graph diagnostics, semantic reward, and velocity-level action abstraction.

**Table S9 Ablation summary over the last 20% evaluation points.**

| Variant | Seeds | Seed IDs | Return | Tail dist. | Lost | Sat. | Control |
|---|---|---|---|---|---|---|---|
| MAPPO-raw-tau6 | 3 | 0,1,2 | 233.06 +/- 17.00 | 0.136 +/- 0.107 | 0.018 +/- 0.022 | 0.000 +/- 0.000 | 0.029 +/- 0.004 |
| MAPPO-velocity3 | 3 | 0,1,2 | 212.15 +/- 18.96 | 0.208 +/- 0.060 | 0.000 +/- 0.000 | 0.000 +/- 0.000 | 0.047 +/- 0.002 |
| MAPPO-semantic-state | 3 | 0,1,2 | 205.48 +/- 10.04 | 0.215 +/- 0.029 | 0.000 +/- 0.000 | 0.000 +/- 0.000 | 0.039 +/- 0.003 |
| STG-MAPPO-full | 3 | 0,1,2 | 254.81 +/- 11.66 | 0.014 +/- 0.003 | 0.000 +/- 0.000 | 0.000 +/- 0.000 | 0.014 +/- 0.006 |

The Table S9 indicate that velocity abstraction alone is insufficient to reproduce the full STG-MAPPO performance. Semantic state without semantic reward also does not fully exploit task-phase information. The strongest behavior appears when semantic state, semantic graph diagnostics, semantic reward, and velocity-level execution are used together.

# F. Pseudocode

**Algorithm S1 STG-MAPPO training with semantic task graph.**

```
Input: AUV6DOF environment, semantic feature builder, PPO actor-critic, horizon T.
```

```
Initialize actor parameters theta, critic parameters phi, and rollout buffer D.
for each training iteration do
    Reset parallel environments under the selected curriculum stage.
    for t = 1,...,T do
        For each AUV i, construct normalized local observation o_i.
        Build semantic phase, observation-quality, communication, and role features.
        Build semantic task-graph diagnostics from neighbor and target-tracking states.
        Actor pi_theta outputs normalized velocity action a_i in [-1,1]^3.
        Map a_i to desired velocity and then to six-degree-of-freedom generalized force tau_i.
        Step AUV dynamics, target motion, reward decomposition, done, and info diagnostics.
        Store transition and reward terms in D.
    end for
    Estimate advantages and update actor/critic with clipped PPO objective.
    Periodically evaluate and save the best checkpoint.
end for
```

**Algorithm S2 Best-checkpoint stress-test evaluation.**

```
Input: completed runs, algorithms, seeds, stress conditions, K evaluation episodes.
for each scenario in {medium, hard} do
    for each algorithm and seed do
        Load ckpt_best.pth.tar from the completed training run.
        for each stress condition c do
            Apply condition-specific target-speed, initialization, sensing, or communication override.
            Run K independent evaluation episodes with horizon 500.
            Record return, tail100 distance, tracking error, target-lost rate, action, control, and
communication metrics.
        end for
    end for
    Aggregate mean +/- standard deviation and export paper tables.
end for
```